\documentclass[letterpaper, preprint, paper,11pt]{AAS}	

\usepackage{bm}
\usepackage{amsmath}
\usepackage{amssymb}

\usepackage[colorlinks=true, pdfstartview=FitV, linkcolor=black, citecolor= black, urlcolor= black]{hyperref}
\usepackage{overcite}
\usepackage{footnpag}			      	
\usepackage{textcomp}

\usepackage{adjustbox}
\usepackage{subcaption}
\usepackage{booktabs}
\usepackage{array}

\usepackage{xcolor}

\definecolor{mplLime}{HTML}{00FF00}
\definecolor{mplGreen}{HTML}{008000}
\definecolor{mplYellow}{HTML}{FFFF00}
\definecolor{mplGold}{HTML}{FFD700}
\definecolor{mplOrange}{HTML}{FFA500}

\usepackage{xcolor}
\usepackage{tikz}
\usetikzlibrary{shapes,arrows,calc,patterns,angles,quotes,fadings,decorations.pathreplacing}
\usepackage{tikz-3dplot}
\usepackage{tkz-graph}
\tikzstyle{vertex}=[circle, draw, inner sep=0pt, minimum size=8pt]

\tikzset{%
  >=latex, 
  inner sep=0pt,%
  outer sep=1pt,%
  mark coordinate/.style={inner sep=0pt,outer sep=0pt,minimum size=3pt,
    fill=black,circle}%
}

\newcommand{\ra}[1]{\renewcommand{\arraystretch}{#1}} 

\PaperNumber{26-675}

\begin{document}

\title{Principal Component Analysis for \\
Lunar Crater Detection}

\author{Travis Driver\thanks{Research Engineer II, School of Aerospace Engineering, Georgia Institute of Technology, Atlanta, GA 30332.}  
\ and John A. Christian\thanks{Associate Professor, School of Aerospace Engineering, Georgia Institute of Technology, Atlanta, GA 30332.}
}

\maketitle{}

\begin{abstract}
Optical navigation is a critical component for lunar orbiter and lander missions. 
Image-based crater identification has emerged as a promising technology for optical navigation due to the abundance of craters on the lunar surface and the availability of extensive crater catalogs. 
Moreover, due to the relative morphological homogeneity among lunar craters, template matching has been identified as a promising approach for identification.
In this paper, we propose \textsc{EigenCrater}, an \textit{automated} crater template generation method based on principal component analysis of crater digital elevation maps (DEMs). 
We demonstrate superior detection and position estimation performance relative to \textit{hand-picked} templates on simulated lunar imagery. 
\end{abstract}

\section{Introduction}

Optical navigation is a critical enabling technology for autonomous spacecraft operations in cislunar space, supporting tasks such as orbit determination, descent guidance, and precision landing in GPS-denied environments. 
As lunar missions increasingly shift toward higher levels of onboard autonomy, vision-based navigation systems play a central role in reducing operational risk, limiting reliance on ground-in-the-loop decision making, and lowering mission costs. 
By exploiting visual information from onboard cameras, optical navigation allows the spacecraft to estimate its motion either relative to previously mapped surface features or directly from features extracted online, enabling robust navigation without external positioning infrastructure.

Among the many visual cues available on the lunar surface, impact craters have emerged as particularly attractive landmarks for optical navigation. 
Craters are abundant, persistent, and globally distributed, and extensive lunar crater catalogs derived from orbital imagery and laser altimetry are readily available~\cite{robbins2019catalog, wang2021lu1319373, lagrassa2025lu5m812tgt}. 
As a result, image-based crater identification has been widely studied as a navigation modality and has already been deployed on recent lunar missions. 
Indeed, crater identification has been used for optical navigation on a variety of lunar missions, including Intuitive Machines’ IM-2~\cite{thrasher2024template} and JAXA’s SLIM~\cite{ishida2025slim_vbn}. 
These missions relied on template matching approaches, in which representative crater templates are stored onboard and correlated with incoming imagery to generate crater detections and associated navigation measurements.

Template matching is particularly well-suited to crater detection due to the relative morphological homogeneity of lunar craters across a wide range of scales. 
However, a key challenge in these systems lies in the construction of effective crater templates. 
Existing approaches often rely on hand-designed or manually selected templates, which may fail to capture the diversity of crater morphologies. 
JAXA’s SLIM mission addressed this limitation by leveraging templates formed from the principal components of crater image patches extracted from Apollo, LRO, and SELENE imagery. 
While effective, such image-space approaches entangle crater geometry with illumination and sensor-specific effects, limiting their generalizability across viewing and lighting conditions.

In this work, we propose \textbf{\textsc{EigenCrater}}, an automated crater template generation framework that operates directly in the space of crater digital elevation maps (DEMs), rather than intensity imagery. 
The etymology of the name draws inspiration from the seminal EigenFaces~\cite{turk1991eigenfaces} method for face recognition. 
We perform principal component analysis over a large corpus of local crater elevation (and optionally albedo) maps, yielding a compact, low-dimensional representation of crater geometry. 
Craters are then clustered in this reduced space, and the mean of each cluster is used to define a representative crater template. 
These templates can be rendered into the image domain using an appropriate reflectance model and known or estimated sun vector information, enabling their use within standard template matching pipelines.

We evaluate the proposed approach on simulated lunar imagery and demonstrate that \textsc{EigenCrater} yields superior crater detection and position estimation performance compared to \textit{hand-picked} templates~\cite{thrasher2024template}. 
By decoupling crater shape from illumination effects and automating template construction, our method produces physically meaningful, diverse templates that better span the space of observable crater appearances. 
These results suggest that \textit{data-driven template generation} is a promising direction for robust and scalable crater-based optical navigation systems for future lunar missions.
\section{Related Work}

\subsection{Optical Navigation and Vision-Based State Estimation}

Vision-based navigation has long been studied as a means of enabling autonomous spacecraft operation in GPS-denied environments. \cite{christian2026} 
Broadly, optical navigation methods can be divided into \textit{relative} navigation approaches---such as visual odometry (VO) and visual simultaneous localization and mapping (VSLAM)---and \textit{absolute} navigation approaches that localize a vehicle with respect to a known surface map or landmark catalog. 
Feature-based VO and VSLAM systems that use keypoint detectors and descriptors such as SIFT~\cite{lowe2004distinctive} or ORB~\cite{rublee2011orb} estimate motion by tracking image features over time and have demonstrated success in terrestrial and space robotics contexts~\cite{dor2021vslam,dor2024astroslam,christian2021vo}. 
However, these methods provide \textit{relative} pose estimates and are subject to drift unless augmented with absolute measurements or a robust loop closure framework.

\textit{Absolute} optical navigation techniques, instead, take advantage of prior knowledge of the environment, often in the form of mapped landmarks, to provide drift-free state updates. 
On the Moon, the availability of high-resolution global maps and crater catalogs has made crater-based navigation a particularly attractive option for absolute pose estimation, especially during descent and landing phases, where absolute localization is critical.

\subsection{Template Matching}

Template matching has emerged as a natural approach for crater detection due to the relative morphological homogeneity of lunar craters. 
Methods have traditionally relied on hand-designed or manually selected crater templates that are correlated with incoming imagery to identify candidate detections~\cite{grumpe2013generative, thrasher2024template}. 
While computationally efficient and well-suited to onboard implementation, such templates often fail to capture the diversity of crater shapes arising from differences in type (i.e., simple, transitional, or complex~\cite{kruger2018deriving}), degradation state, and local topography.

To address these limitations, data-driven template construction techniques have been explored. 
Notably, JAXA's SLIM leveraged principal component analysis (PCA) over crater image patches extracted from multiple lunar datasets to construct a compact basis of crater appearance~\cite{ishida2025slim_vbn}. 
While effective in practice, such image-space approaches inherently couple crater geometry with illumination and sensor-specific effects, potentially limiting robustness across lighting conditions and mission profiles. 
Instead, we leverage statistical machine learning techniques, namely PCA and $k$-means clustering, to generate crater elevation templates directly from global DEMs, thereby capturing intrinsic crater geometry independent of imaging conditions.

\subsection{Edge-based Approaches}

Early approaches to crater detection focused primarily on identifying elliptical edges and characteristic shadow--highlight patterns produced by impact craters. 
Flores et al.~\cite{flores2003crater} combined edge detection with the circular Hough transform to extract crater rims from monocular imagery. 
However, this approach is effective only under near-nadir viewing geometry and favorable illumination conditions. 
Cheng et al.~\cite{cheng2003optical} extended this paradigm by introducing a multi-stage pipeline that applied Canny edge detection with dual kernel sizes---small kernels to capture sharp crater rims and larger kernels to detect more diffuse boundaries---followed by illumination-informed grouping of shadow and highlight profiles. 
Despite these refinements, shadow and highlight extraction remains highly sensitive to illumination geometry and the selection of shadow and highlight thresholds, limiting robustness under realistic mission conditions~\cite{hamel2025crater}.

\subsection{Deep Learning}

Methods based on deep learning have been shown to improve robustness to variations in illumination and viewing geometry when trained on sufficiently representative datasets. 
LunaNet~\cite{downes2021lunanet} leveraged a U-Net architecture to segment crater rims from lunar imagery, followed by ellipse fitting to recover crater geometry. 
Reference~\citenum{delatte2019segmentation} conducted an extensive study of U-Net variants and segmentation strategies for crater identification in Martian imagery. 
Object detection frameworks such as Mask Region-Based Convolutional Neural Networks (Mask R-CNNs)~\cite{he2017maskrcnn} have also been widely applied to crater detection by generating bounding boxes and segmentation masks~\cite{catalan2021rcnn,mclaughlin2022rcnn,catalan2025rcnn}, as well as YOLO-based architectures~\cite{redmon2016you,mu2023yolo,la2023yololens,kilduff2025yolo}. 

Ellipse R-CNN~\cite{dong2021ellipse} augments the Mask R-CNN framework with an ellipse regression head, enabling direct prediction of ellipse parameters for elliptical objects. 
While originally proposed for applications such as face detection, Ellipse R-CNN was subsequently applied to lunar crater detection by Reference~\citenum{doppenberg2021ercnn}. 
This formulation is particularly attractive for crater-based navigation, as it eliminates the need for an explicit ellipse-fitting step applied to segmentation masks or bounding boxes. 
Reference~\citenum{doppenberg2021ercnn} released both code and pretrained model weights, which we use in this work to compare the pretrained model and a version fine-tuned on imagery generated using NASA’s Vira renderer~\cite{gnam2025vira} against the proposed method.

\section{Principal Component Analysis for Crater Elevation Templates}
\label{sec:pca_craters}

We leverage principal component analysis (PCA) to find the most salient geometric variations in local crater digital elevation maps (DEMs) across the lunar surface. 
The principal components can then be used to cluster local crater DEMs in a lower dimensional space, and the population mean of these clusters may be used as a representative template for image crater detection. 


\subsection{Local Crater DEM Extraction} \label{sec:dem_extraction}

Let $\{\mathbf{E}_i\}_{i=1}^{N}$ denote a collection of $N$ crater-centered elevation map patches, where each patch $\mathbf{E}_i \in \mathbb{R}^{H \times W}$ is extracted from a global lunar DEM and resampled to a fixed spatial resolution and extent. 
Each elevation patch is vectorized into a column vector
\begin{equation}
    \mathbf{x}_i = \mathrm{vec}(\mathbf{E}_i) \in \mathbb{R}^{D},
\end{equation}
where $D = H \times W$. Stacking all samples yields the data matrix
\begin{equation}
    \mathbf{X} = 
    \begin{bmatrix}
    \mathbf{x}_1^\top \\
    \mathbf{x}_2^\top \\
    \vdots \\
    \mathbf{x}_N^\top
    \end{bmatrix}
    \in \mathbb{R}^{N \times D}.
\end{equation}
Prior to PCA, each vectorized elevation map is mean-centered by subtracting the empirical mean elevation vector
\begin{equation}
    \bar{\mathbf{x}} = \frac{1}{N} \sum_{i=1}^{N} \mathbf{x}_i,
\end{equation}
resulting in the centered data matrix $\tilde{\mathbf{X}}$.

\begin{figure}[tb!]
    \centering
    \begin{subfigure}{.49\linewidth}
        \includegraphics[width=\linewidth]{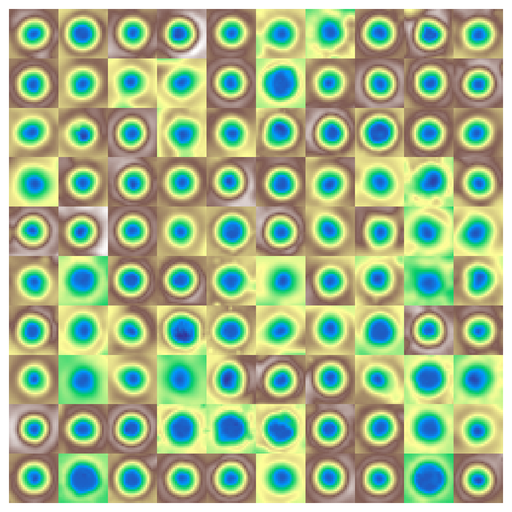}
    \end{subfigure}
    \hfill
    \begin{subfigure}{.49\linewidth}
        \includegraphics[width=\linewidth]{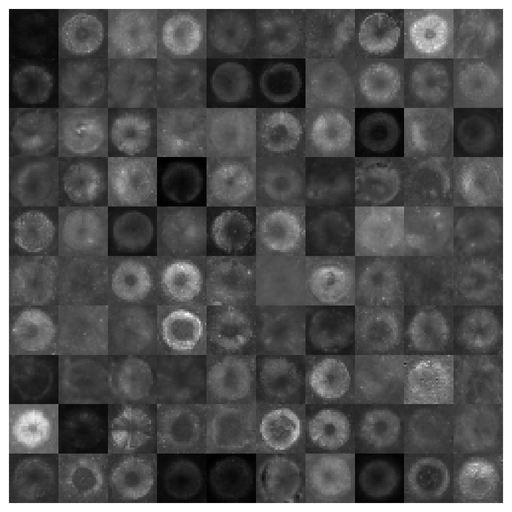}
    \end{subfigure}
    \caption{\textbf{Example extracted elevation (left) and albedo (right) maps.}}
    \label{fig:example-elevations-albedos}
\end{figure}

We leverage the Robbins crater database~\cite{robbins2019catalog} and filter to craters with a radius between 2 and 16 km and an eccentricity of $\leq 0.3$. 
The Robbins parameters are then used to extract a physically square local DEM at $1.2\times$ the Robbins circular radius from the near-global 100 m/pixel digital terrain model (GLD100) derived from stereo image data acquired by the Wide-angle Camera (WAC) of the Lunar Reconnaissance Orbiter Camera (LROC) system~\cite{scholten2012gld100}, which are then resized to $25\times 25$ DEMs. 
Symmetry of the extrated craters is constrained by rotating the DEM in three 90$^\circ$ increments and ensuring that the maximum absolute difference between the original and rotated template is $<40\%$ of the depth of the crater, where depth is measured from the center of the crater to the highest point on the rim. 
Craters with a depth-to-diameter ratio of $<0.1$ are also filtered. 
This filtering process resulted in 1780 local DEMs. 

The original and three rotated DEMs are stored for the PCA step (discussed in the next subsection) to account for any lingering asymmetries. 
An equivalent process may be conducted on the LROC WAC empirically normalized mosaics~\cite{boyd2012empirical} to extract a measure of albedo over the same topography. 
Representative local elevation and albedo maps extracted using this process are shown in Fig.~\ref{fig:example-elevations-albedos}.


\subsection{Principal Component Analysis} \label{sec:pca}

PCA seeks an orthonormal basis that captures the directions of maximum variance in the dataset. 
This is achieved by computing the eigendecomposition of the empirical covariance matrix
\begin{equation}
\mathbf{C} = \frac{1}{N-1} \tilde{\mathbf{X}}^\top \tilde{\mathbf{X}} \in \mathbb{R}^{D \times D}.
\end{equation}
Let $\{\lambda_k, \mathbf{v}_k\}_{k=1}^{D}$ denote the eigenvalue--eigenvector pairs of $\mathbf{C}$, sorted in descending order of eigenvalue magnitude. 
The eigenvectors $\mathbf{v}_k$ define the \textit{principal components}, each corresponding to a dominant mode of variation in crater elevation structure.

In practice, only the first $K \ll D$ principal components are retained, forming the projection matrix
\begin{equation}
\mathbf{V}_K = \begin{bmatrix} \mathbf{v}_1 & \mathbf{v}_2 & \cdots & \mathbf{v}_K \end{bmatrix}.
\end{equation}
Each crater elevation map is then projected into the reduced-dimensional subspace as
\begin{equation}
\mathbf{z}_i = \mathbf{V}_K^\top (\mathbf{x}_i - \bar{\mathbf{x}}) \in \mathbb{R}^{K}.
\end{equation}
This representation preserves the most salient geometric variations among crater elevation profiles while substantially reducing dimensionality.

\begin{figure}[tb!]
    \centering
    \begin{subfigure}{0.32\linewidth}
        \includegraphics[height=.85\linewidth]{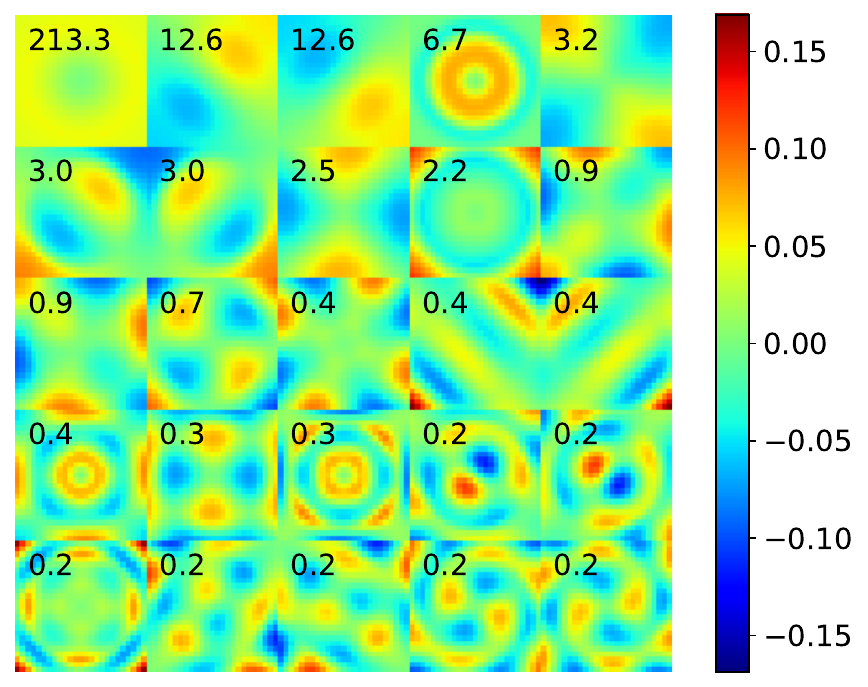}
        \caption{Elevation only}
    \end{subfigure}%
    \hfill
    \begin{subfigure}{0.64\linewidth}
        \begin{subfigure}{0.5\linewidth}
            \includegraphics[height=.85\linewidth]{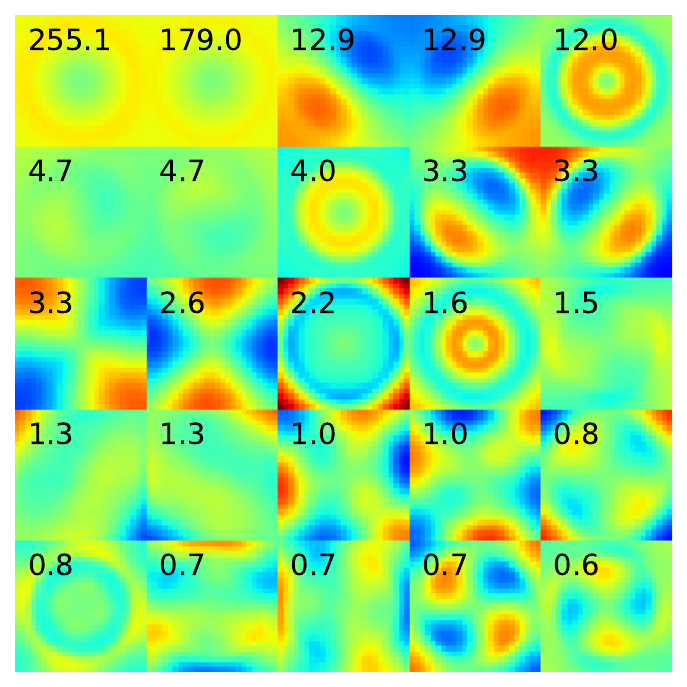}
        \end{subfigure}%
        \hspace{-20pt}
        \begin{subfigure}{0.5\linewidth}
            \includegraphics[height=.85\linewidth]{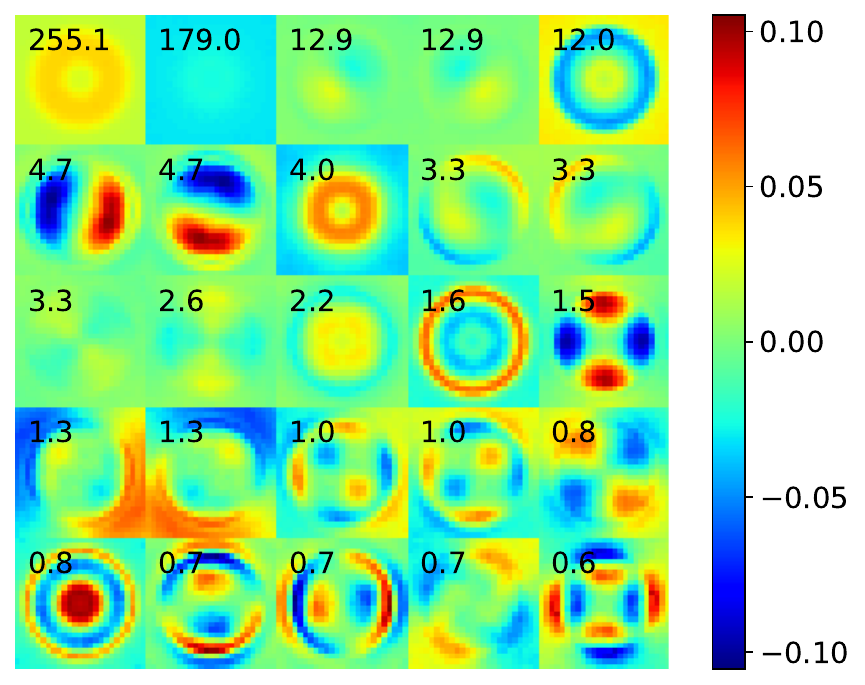}
        \end{subfigure}%
        \caption{Elevation (left) \& albedo (right)}
    \end{subfigure}
    \caption{\textbf{Tiled top 25 computed principal components for elevation data (left) and elevation and albedo data (right).} The associated eigenvalue for each principal component is provided in the top left of each principal component tile.}
    \label{fig:tiled-principal-components}
\end{figure}

The principal components computed over the extracted local DEMs (and albedo maps) are shown Figure \ref{fig:tiled-principal-components}. 
In the context of lunar craters, the leading principal components correspond to physically meaningful variations, such as overall crater depth, rim-to-floor contrast, rim asymmetries, and secondary topographic features within the crater interior. 
By operating in this reduced space, craters with similar morphological characteristics are mapped to nearby points, facilitating robust clustering.
The low-dimensional PCA embeddings $\{\mathbf{z}_i\}$ are used as input to $k$-means clustering~\cite{arthur2007kmeans} to group craters with similar elevation structure, where we consider the first $K=25$ principal components. 

Cluster centroids are computed in the PCA space and then mapped back to the original elevation domain via the inverse transform, yielding representative crater elevation templates. These templates capture the dominant structural modes of crater morphology present in the dataset and are subsequently used for template-based crater identification.
When considering both elevation and albedo information, a scale factor of 20 is applied to the albedos to bring the total variance in line with that of the elevation to ensure equal weighting of both features. 


\subsection{Crater Template Rendering \& Matching} \label{sec:rendering}

\begin{figure}[t]
    \centering
    \tdplotsetmaincoords{70}{120}
\begin{tikzpicture}[tdplot_main_coords, scale=1.5, every node/.style={scale=1.3}]

    \coordinate (oo) at (0,0,0);
    \coordinate (bl) at (-1/2,-1/2,0);
    \coordinate (br) at (1/2,-1/2,0);
    \coordinate (tl) at (-1/2,1/2,0);
    \coordinate (tr) at (1/2,1/2,0);
    \filldraw[very thick, fill=gray!40] (tl) -- (tr) -- (br) -- (bl) -- cycle;
    \node[label=west:{$\mathbf{v}\,$}] (a) at (0,0,0) {};

    \coordinate (ss) at (3/0.75, 0, 4/0.75);
    \coordinate (ss2) at (3/0.8, 0, 4/0.8);
    
    \coordinate (cc) at (0, 3/0.80, 4/0.80);
    \node[label=north east:{}] () at (cc) {};
    
    \coordinate (cp) at (0, 3/1.05, 4/1.05);
    \coordinate (ctl) at (0.5, 2.5, 4.5);
    \coordinate (ctr) at (0.5, 3.5, 4.5);
    \coordinate (cbl) at (0, 2.5, 3.5);
    \coordinate (cbr) at (0, 3.5, 3.5);
    \draw[very thick,gray] (ctl) -- (ctr);
    \draw[very thick,gray] (ctl) -- (cbl);
    \draw[very thick,gray] (cbl) -- (cbr);
    \draw[very thick,gray] (ctr) -- (cbr);
    
    \filldraw[very thick, fill=gray!20] (ctl) -- (ctr) -- (cbr) -- (cbl) -- cycle;
    \draw[very thick] (cc) -- (ctl);
    \draw[very thick] (cc) -- (ctr);
    \draw[very thick] (cc) -- (cbr);
    \draw[very thick] (cc) -- (cbl);

    \shadedraw[opacity = .5] (3/0.75, 0, 4/0.75) circle (.2cm);
    \fill[black] (oo) circle (0.10);
    \filldraw[draw=none, fill=gray!90] (cp) circle (.05cm);
    \filldraw[thick, fill=black] (cc) circle (.05cm);
    
    \node[inner sep=0pt] (camera) at (0, 3/1.1, 4/1.1) {};
    \node[inner sep=0pt] (tmp) at (cp) {};

    \draw[dashed] (ss2) -- (oo);
    \draw[dashed] (cp) -- (oo);
    \draw[very thick,->] (oo) -- (0,0,5/1.7) coordinate (nhat) node [black,right] {$\,\mathbf{n}$};
    \draw[very thick,->] (oo) -- (3/1.5,0,4/1.5) coordinate (ihat) node [black,left] {$\mathbf{s}\,$};
    \draw[very thick,->] (oo) -- (0,3/1.7,4/1.7) coordinate (ehat) node [black,right] {$\,\mathbf{e}$};
    
    \tdplotsetrotatedcoords{0}{-90}{0}
    \tdplotdrawarc[tdplot_rotated_coords]{(0,0,0)}{1.8}{0}{90-53.13}{anchor=south}{$\quad\varepsilon$}
    \tdplotsetrotatedcoords{90}{-90}{0}
    \tdplotdrawarc[tdplot_rotated_coords]{(0,0,0)}{1.8}{0}{-36.87}{anchor=south}{$\iota\quad$}
    \tdplotsetrotatedcoords{45}{-62.0616}{0}
    \tdplotdrawarc[tdplot_rotated_coords]{(0,0,0)}{1.8}{24}{-24}{anchor=north}{$\phi$}
\end{tikzpicture}
    \vspace{-20pt}
    \caption{\textbf{Photometry conventions.}}
    \label{fig:angles}
\end{figure}
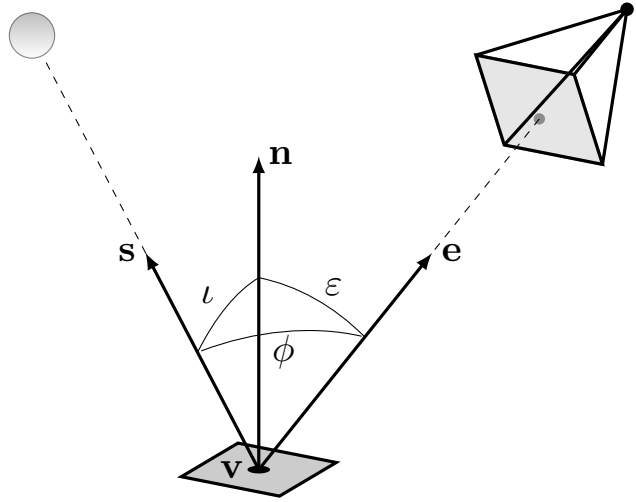

The crater template used for matching is rendered using a Lunar-Lambert reflectance model. 
The $25\times 25$ DEM is first converted to a triangular mesh, where the DEM pixel centers are treated as vertices. 
The facet normals are computed for the resulting mesh via 
\begin{equation}
    \mathbf{n}_\mathrm{facet} = (\mathbf{v}_1 - \mathbf{v}_0) \times (\mathbf{v}_2 - \mathbf{v}_0), 
\end{equation}
where $\mathbf{v}_i$ are the vertices of the triangular facet. 
The surface (vertex) normals $\mathbf{n}$ are then computed as a corner angle-weighted average of the adjacent face normals~\cite{thurrner1998computing}.

Next, the computed surface normals may be combined with the Sun vector $\mathbf{s}$ and the emission vector $\mathbf{e}$ using the Lunar-Lambertian reflectance model (or, more formally, the \textit{bidirectional radiance factor})~\cite{mcewen1991,mcewen1996precise} to generate a rendering of the computed mesh:
\begin{equation}
    r(a, \iota, \varepsilon, \phi) = a \left[(1-g(\phi)) \cos\iota + g(\phi)\frac{2\cos\iota}{\cos\iota + \cos\varepsilon}\right], 
\end{equation}
where $a$ is the surface albedo, $\iota$ is the angle between the incoming light and the surface normal, or the \textit{incidence angle}, $\varepsilon$ is the angle between the emitted light and the surface normal, or the \textit{emission angle}, and $\phi$ is the angle between the emitted light and the incoming light, or the \textit{phase angle} (see Fig. \ref{fig:angles}). 
For the elevation only case, the albedo $a$ is set to a uniform constant over the entire template. 
The function $g$ weights the contribution of the Lambertian and Lommel-Seeliger photometric functions according to the phase angle, where we leverage the exponential form proposed in Reference~\citenum{gaskell2008}:
\begin{equation}
    g(\phi) = \exp\left(\frac{-\phi}{60}\right), 
\end{equation}
where $\phi$ is in degrees. For additional information on reflectance modeling, see Refs.~\citen{christian2026} and \citen{shepard2017}.

We treat the Sun as a point source delivering collimated light to the surface owing to the large distance to the Sun (i.e., the Sun subtends an angle of $0.5^\circ$ at Earth~\cite{shepard2017}) and the lack of a lunar atmosphere to scatter the incoming light. 
Thus, all vertices share the same Sun vector. 
Shadows are computed by tracing rays from each vertex in the Sun vector direction and recording which rays intersect the mesh, where we leverage the Embree~\cite{wald2014embree} ray tracer used in the Trimesh~\cite{trimesh} library.

The template is rendered once and matched against an image pyramid at scales of $1\times$, $1/2\times$, and $1/4\times$ the original image dimensions using normalized cross correlation (NCC). 
Detections across templates and scales are filtered using a correlation threshold of 0.7 and non-maximum suppression with an overlap threshold of 0.4, and the top 30 detections with respect to the correlation score are kept for the matching step. 
Subpixel localization is achieved by fitting a second-order polynomial to a $5\times 5$ neighborhood centered at the peak of the correlation surface and solving for its stationary point~\cite{tong2019image}, analogous to the extremum interpolation used in SIFT~\cite{lowe2004distinctive}. A very similar approach, with a $3 \times 3$ neighborhood, was used for subpixel localization of SPC templates in the OSIRIS-REx Natural Feature Tracking (NFT) algorithm.\cite{olds2015}


\subsection{Homographic Adaptation} \label{sec:homography}

The local template frame and the camera frame are assumed to be aligned when considering nadir-pointing imagery. 
However, this will not be the case in general, and it is beneficial to apply a transformation to the image to better compensate for off-nadir viewing. 
To accomplish this, we will compute a homography\cite{christian2026} between the camera frame and a nadir-point frame, and warp the image according to this transformation. 

We assume knowledge of an estimate of the attitude $\mathbf{R}_\mathcal{CM}$ of the camera frame $\mathcal{C}$ with respect to a Moon-centered, Moon-fixed (MCMF) frame $\mathcal{M}$ and the relative position $\mathbf{r}_\mathrm{CM}^\mathcal{M}$, which points from the origin of the MCMF frame $\mathrm{M}$ to the origin of the camera frame $\mathrm{C}$, expressed in the MCMF frame. 
Recall that the rows of $\mathbf{R}_\mathcal{CM}$ represent the basis vectors of $\mathcal{C}$ expressed in $\mathcal{M}$:
\begin{equation}
    \mathbf{R}_\mathcal{CM} = 
    \begin{bmatrix}
        \left(\mathbf{c}_1^\mathcal{M}\right)^\top \\
        \left(\mathbf{c}_2^\mathcal{M}\right)^\top \\
        \left(\mathbf{c}_3^\mathcal{M}\right)^\top \\
    \end{bmatrix}
\end{equation}
Thus, we may solve for the distance $d_\mathrm{surface}$ from the camera origin to the surface in the boresight direction $\mathbf{c}_3^\mathcal{M}$ by solving the quadratic\footnote{See Exercise 6.12 in Ref.~\citen{christian2026} for details.}
\begin{align}
    \|\mathbf{r}_\mathrm{CM}^\mathcal{M} + d_\mathrm{surface}\, \mathbf{c}_3^\mathcal{M}\|^2 = R_\mathrm{Moon}^2 \Leftrightarrow 
    d_\mathrm{surface}^2 + \left[2\left(\mathbf{r}_\mathrm{CM}^\mathcal{M}\right)^\top\mathbf{c}_3^\mathcal{M}\right]d_\mathrm{surface} + \left(\|\mathbf{r}_\mathrm{CM}^\mathcal{M}\|^2 - R_\mathrm{Moon}^2\right) = 0,
\end{align}
where we've use the fact that $\|\mathbf{c}_3^\mathcal{M}\| = 1$. 
This gives
\begin{equation}
    d_\mathrm{surface} = -\left(\mathbf{r}_\mathrm{CM}^\mathcal{M}\right)^\top\mathbf{c}_3^\mathcal{M} \pm \sqrt{\left[\left(\mathbf{r}_\mathrm{CM}^\mathcal{M}\right)^\top\mathbf{c}_3^\mathcal{M}\right]^2 - \left(\|\mathbf{r}_\mathrm{CM}^\mathcal{M}\|^2 - R_\mathrm{Moon}^2\right)},
\end{equation}
where we choose the smaller solution, as the larger solution puts the intersection point on the other side of the Moon. 

\begin{figure}[tb!]
    \centering
    \includegraphics[width=0.6\linewidth]{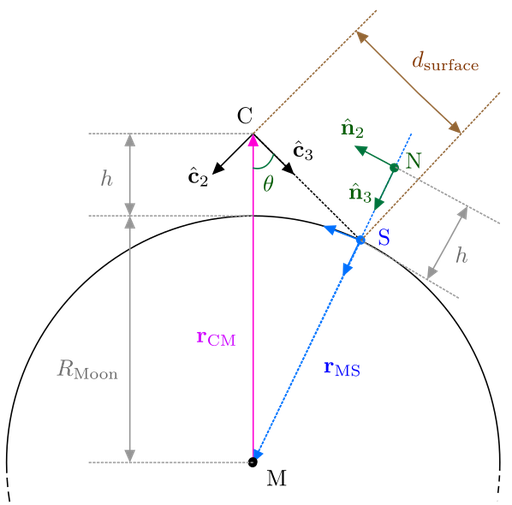}
    \caption{\textbf{Homography adaptation conventions.}}
    \label{fig:placeholder}
\end{figure}

Thus, we may compute the vector pointing from the camera's origin to the intersection point between the cameras boresight and the mean lunar surface as
\begin{equation}
    \mathbf{r}_\mathrm{SM}^\mathcal{M} = \underbrace{\mathbf{c}_3^\mathcal{M}\, d_\mathrm{surface}}_{\mathbf{r}_\mathrm{SC}^\mathcal{M}} + \underbrace{\mathbf{R}_\mathcal{CM}^\top\left( -\mathbf{r}_\mathrm{MC}^\mathcal{C} \right)}_{\mathbf{r}_\mathrm{CM}^\mathcal{M}}. 
\end{equation}
We can then scale this vector to get a vector at the same altitude $h$ as the original camera frame to get the origin of the nadir frame $\mathcal{N}$ via 
\begin{equation}
    \mathbf{r}_\mathrm{NM}^\mathcal{M} = \mathbf{r}_\mathrm{SM}^\mathcal{M}\, \frac{R_\mathrm{Moon} + h}{R_\mathrm{Moon}} = \mathbf{r}_\mathrm{SM}^\mathcal{M}\, \frac{\|\mathbf{r}_\mathrm{CM}\|}{R_\mathrm{Moon}}. 
\end{equation}
Finally, we compute the basis vectors $\mathbf{n}_i$ of $\mathcal{N}$ as 
\begin{align}
    \mathbf{n}_3^\mathcal{M} &= -\mathbf{r}_\mathrm{NM}^\mathcal{M} / \|\mathbf{r}_\mathrm{NM}^\mathcal{M}\|, \\
    \mathbf{n}_1^\mathcal{M} &= \mathbf{c}_2^\mathcal{M} \times \mathbf{n}_3^\mathcal{M}, \\
    \mathbf{n}_2^\mathcal{M} &= \mathbf{n}_3^\mathcal{M} \times \mathbf{n}_1^\mathcal{M}, \\
\end{align}
yielding the rotation
\begin{equation}
    \mathbf{R}_\mathcal{NM} = 
    \begin{bmatrix}
        \left(\mathbf{n}_1^\mathcal{M}\right)^\top \\
        \left(\mathbf{n}_2^\mathcal{M}\right)^\top \\
        \left(\mathbf{n}_3^\mathcal{M}\right)^\top \\
    \end{bmatrix}.
\end{equation}

\begin{figure}[tb!]
    \centering
    \begin{subfigure}{.49\linewidth}
        \includegraphics[width=\linewidth]{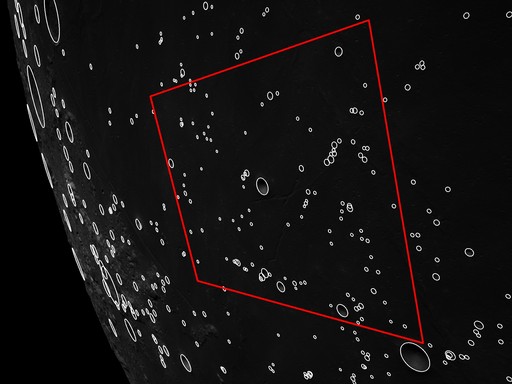}
    \end{subfigure}%
    \hfill
    \begin{subfigure}{.49\linewidth}
        \includegraphics[width=\linewidth]{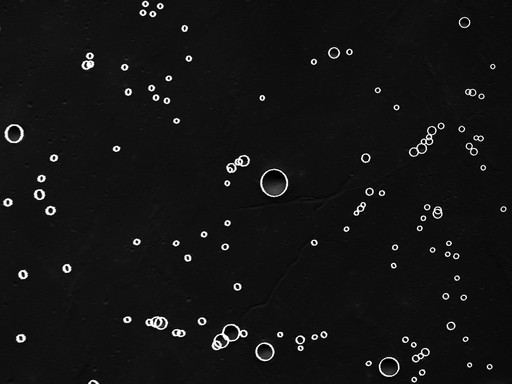}
    \end{subfigure}%
    \caption{\textbf{Illustration of homographic adaptation process.} The original image (with craters outlined in white) is warped according to the computed homography.}
    \label{fig:img-warped}
\end{figure}

Next, let $\mathbf{s}_i$ be the columns of $\mathbf{R}_\mathcal{CS} = \mathbf{R}_\mathcal{CM}\mathbf{R}_\mathcal{NM}^\top$ and $\mathbf{r}_\mathrm{PC}^\mathcal{C} =\mathbf{R}_\mathcal{CM}\mathbf{r}_\mathrm{PC}^\mathcal{M}$. 
Then, the homography between the camera and surface frame $\mathcal{S}$ is computed as 
\begin{equation}
    \mathbf{H}_\mathcal{CS} = \mathbf{K}
    \begin{bmatrix}
        \mathbf{s}_1 & \mathbf{s}_2 & \mathbf{r}_\mathrm{SC}^\mathcal{C} \\
    \end{bmatrix},
\end{equation}
The $3 \times 3$ matrix $\mathbf{K}$ is the camera calibration matrix\cite{christian2026}
\begin{equation}
\mathbf{K} = 
    \begin{bmatrix}
    d_x & 0 & u_p \\
    0 & d_y & v_p \\
    0 & 0 & 1
    \end{bmatrix}
\end{equation}
where $d_x$ and $d_y$ are the ratio of focal lengths to pixel pitch in the $x$- and $y$-directions of the camera frame, and $(u_p, v_p)$ is the principal point of the camera. We assume no detector skewness.

Moreover, the homography from the surface to nadir frame is 
\begin{equation}
    \mathbf{H}_\mathcal{NS} = \mathbf{K}
    \begin{bmatrix}
        \mathbf{i}_1 & \mathbf{i}_2 & \mathbf{r}_\mathrm{SN}^\mathcal{N} \\
    \end{bmatrix},
\end{equation}
where $\mathbf{i}_i$ is the $i$th column of the identity matrix since the $\mathcal{S}$ and $\mathcal{N}$ frames are axially aligned, i.e., $\mathbf{R}_\mathcal{SN} = \mathbf{R}_\mathcal{NS} = \mathbf{I}_{3\times 3}$. 
Then, the homography from the camera to the nadir frame is 
\begin{equation}
    \mathbf{H}_\mathcal{NC} = \mathbf{H}_\mathcal{NS} \mathbf{H}_\mathcal{CS}^{-1}.
\end{equation}
An example of applying this homographic adaptation is illustrated in Figure~\ref{fig:img-warped}. 

The rotation $\mathbf{R}_\mathcal{NC}$ may be used to transform the emission vector $\mathbf{e}^\mathcal{C} = [0,0,-1]^\top$ and Sun vector $\mathbf{s}^\mathcal{C}$ into the nadir frame before rendering the template, i.e., $\mathbf{e}^\mathcal{N} = \mathbf{R}_\mathcal{NC}\mathbf{e}^\mathcal{C}$ and $\mathbf{s}^\mathcal{N} = \mathbf{R}_\mathcal{NC}\mathbf{s}^\mathcal{C}$. 
Then, the homography may be used to either warp the nadir-rendered template back into the camera frame, or the image may be warped into the nadir frame. 
We found that warping the image rather than the template yielded improved detection performance, which is discussed further in the \nameref{sec:results} section.

\section{Experimental Setup}

\subsection{Image Generation and Catalog Projection}

We evaluate the proposed method on 1000 simulated lunar images with a resolution of $1536\times 2048$ and a field-of-view (FOV) of $45.01^\circ \times 57.85^\circ$ using the Vira rendering library~\cite{gnam2025vira}. 
The images are rendered using the LRO LOLA DEM Coregistered with Selene Data (SLDEM) with the LROC WAC Empirically normalized mosaic as a measure of the albedo. 
The images were rendered at an altitude of 100 km with illumination angles uniformly sampled between 0$^\circ$ and $80^\circ$ and emission angles between $0^\circ$ and $60^\circ$.

Craters from the LU1319373 catalog~\cite{wang2021lu1319373} are projected into the image~\cite{christian2021lunar} to assess the detection performance of the proposed method. 
Only craters within the image bounds and with a projected semi-minor axis length $>$5 pixels and a semi-major axis length of $<$105 pixels are kept for the subsequent identification step. 
The elevation of the craters center with respect to the mean lunar radius was interpolated from a global LOLA DEM using the provided center longitude and latitude coordinates~\cite{andolfo2025neural}, and the final radial distance of each crater in the catalog was further augmented by the crater depth information provided by the LU1319373 catalog.

\subsection{Identification and Position Estimation}

We also test the efficacy of crater detections for position estimation. 
As described previously, the top 30 detections are kept for the subsequent matching step. 
Detections are matched with projected catalog craters using geometric descriptors formed from the interior angles of the triads of the crater centers, where the interior angles are stored counter-clockwise with respect to the largest interior angle~\cite{thrasher2023triangles}. 
The true pose of the camera is corrupted with zero mean Gaussian noise with a standard deviation of 1 km for position and 0.01$^\circ$ for attitude before catalog projection, and only catalog craters with centers within 60 pixels of a detected crater center are used for matching, where $\leq$8000 catalog triads are randomly selected for description. 

The resulting detected and catalog descriptors are matched using a $k$NN matcher with $k=2$ and Lowe's ratio test with a threshold of 0.4~\cite{lowe2004distinctive}. 
The putative descriptor matches are then geometrically verified by triangulating the cameras position using the 2D-to-3D correspondences between the image crater detections and the 3D positions of the catalog craters using the LOST algorithm~\cite{henry2023absolute}. 
The estimated position is then used to reproject the catalog craters, and the projected catalog craters are compared against the detected craters and inliers are registered with a 3 pixel threshold. 
The triad with the most support after 1000 iterations is returned, and the final position is triangulated using the original triad \textit{and} the external inlier craters. 

\subsection{Evaluation Metrics} \label{sec:metrics}

The detection precision, recall, and center errors for each image are computed, and the average of these metrics over all images is reported. 
\textit{Precision} is the ratio of correct to total detections, where a correct detection is a detection within some threshold of a catalog crater, where we use thresholds of 1, 3, 5, and 10 pixels. 
\textit{Recall} is the ratio of correct detections to total \textit{projected} catalog craters (before the center proximity filtering step). 
\textit{Center error} is the average distance between correct detections and their closest catalog crater in the image. 
For assessing position estimation performance, the normalized cumulative error curve for the Euclidean distance between the estimated and true camera position is computed for each image, and the area under the curve (AUC) is reported for thresholds of 0.1, 0.3, 0.5, and 1 km. 
We compute AUC using the explicit integration procedure of Reference \citenum{sarlin2020superglue} rather than coarse histograms. 
We also report the mean, standard deviation, minimum, and maximum of the position errors. 

\subsection{Baselines}

We compare against the template-matching method proposed in Reference~\citenum{thrasher2024template}, which leverages a hand-selected template of the 4.6 km diameter Armstrong crater located at 1.4$^\circ$N, 25.0$^\circ$E. 
We also compare against the deep learning-based method Ellipse R-CNN~\cite{dong2021ellipse}. 
For Ellipse R-CNN, we use open-source pretrained weights from Reference~\citenum{doppenberg2021ercnn} trained on 80,000 images generated using Airbus’ SurRender software~\cite{lebreton2021surrender}, as well as weights fine-tuned on the Vira images with an 80/20 train/test split. 
The 200 held-out test images are used to evaluate both the template-matching and deep learning-based approaches in the following section.
\section{Results} \label{sec:results}


\subsection{Detection Performance}

\begin{table}[tb!]
    \centering
    \ra{1.5}
    \caption{\textbf{Detection statistics.} See the \nameref{sec:metrics} section for metric definitions.}
    \begin{adjustbox}{width=\linewidth}
    \begin{tabular}{lrrrrrrrrrrrrrr}
\toprule
& \multicolumn{4}{c}{Precision [\%]} && \multicolumn{4}{c}{Recall [\%]} && \multicolumn{4}{c}{Center Error [px]} \\
  \cmidrule{2-5} \cmidrule{7-10} \cmidrule{12-15}
 & @1 px & @3 px & @5 px & @10 px && @1 px & @3 px & @5 px & @10 px && @1 px & @3 px & @5 px & @10 px \\ 
\midrule
Ellipse R-CNN~\cite{dong2021ellipse,doppenberg2021ercnn} (pretrained) & 10.42 & 36.44 & 47.44 & 55.88 && 0.85 & 2.74 & 3.35 & 3.77 && 0.65 & 1.56 & 2.18 & 3.02 \\
Ellipse R-CNN~\cite{dong2021ellipse,doppenberg2021ercnn} (finetuned)  & \textbf{18.32} & \textbf{55.87} & \textbf{66.80} & \textbf{74.30} && \textbf{1.40} & \textbf{4.18} & \textbf{4.88} & \textbf{5.33} && \textbf{0.65} & \textbf{1.48} & \textbf{1.86} & \textbf{2.36} \\
\midrule
Armstrong~\cite{thrasher2024template} & 13.16 & 42.04 & 48.83 & 52.79 && 1.04 & 3.12 & \underline{3.48} & 3.66 && \textbf{0.63} & \underline{1.51} & \textbf{1.88} & \textbf{2.38} \\
\textsc{EigenCrater}4                 & \textbf{16.16} & \textbf{52.45} & \textbf{60.49} & \textbf{64.34} && \textbf{1.32} & \textbf{4.09} & \textbf{4.64} & \textbf{4.88} && 0.66 & 1.56 & \underline{1.95} & 2.46 \\
\textsc{EigenCrater}4$+$Albedo        & 12.43 & 38.32 & 43.64 & 47.66 && 1.00 & 2.97 & 3.32 & 3.56 && \textbf{0.63} & 1.52 & 2.03 & 2.95 \\
\textsc{EigenCrater}8                 & \underline{15.96} & \underline{52.21} & \underline{60.42} & \underline{64.24} && \underline{1.30} & \underline{4.07} & \textbf{4.64} & \underline{4.87} && \underline{0.65} & 1.55 & \underline{1.95} & \underline{2.45} \\
\textsc{EigenCrater}8$+$Albedo        & 12.98 & 39.90 & 45.43 & 49.29 && 1.05 & 3.09 & 3.45 & 3.69 && \underline{0.65} & \underline{1.51} & 2.03 & 2.90 \\
\textsc{EigenCrater}16$+$Albedo       & 12.91 & 39.01 & 44.36 & 48.48 && 1.04 & 3.01 & 3.37 & 3.63 && \underline{0.65} & \textbf{1.49} & 2.02 & 2.99 \\
\bottomrule
\end{tabular}

    \end{adjustbox}
    \label{tab:vira-results}
\end{table}

The results in Table~\ref{tab:vira-results} highlight several important trends. 
First, fine-tuning Ellipse R-CNN on imagery generated with NASA’s Vira renderer yields the strongest overall detection performance, achieving the highest precision and recall across all pixel thresholds. 
At a 5 px tolerance, for example, the fine-tuned model attains 66.8\% precision and 4.88\% recall, substantially outperforming the pretrained model, which achieves 47.4\% precision and 3.35\% recall at the same threshold. 
This performance gap underscores the sensitivity of learning-based detectors to domain alignment: despite the \textit{pretrained} Ellipse R-CNN being trained on simulated lunar imagery, it exhibits significantly degraded performance when applied to imagery generated from a different lunar image simulation pipeline. 
Notably, \textsc{EigenCrater}4 and \textsc{EigenCrater}8 outperform the pretrained Ellipse R-CNN across all precision and recall thresholds, indicating that geometry-driven template construction can provide competitive performance even without finetuning. 
These results demonstrate the promise of learning-based crater detection, while simultaneously highlighting the importance of mission-specific training data. 
Note that the seemingly low recall values are due to the fact that recall is computed with respect to \textit{all} projected catalog craters, as mentioned in the \nameref{sec:metrics} section.

\begin{figure*}
    \centering
    \begin{adjustbox}{width=\linewidth}
    \centering
\setlength{\tabcolsep}{-1pt} %
\ra{0}
\begin{tabular}{c>{\centering\arraybackslash}p{3.3cm}>{\centering\arraybackslash}p{3.3cm}>{\centering\arraybackslash}p{3.3cm}>{\centering\arraybackslash}p{3.3cm}}
    & \scriptsize{Armstrong} & \scriptsize{\textsc{EigenCrater}4} & \scriptsize{\textsc{EigenCrater}8} & \scriptsize{\textsc{EigenCrater}8+Albedo} \\
    \parbox[t]{3mm}{\rotatebox[origin=l]{90}{\textcolor{gray}{\tiny{$\epsilon$: 34$^\circ$, $\iota$: 7$^\circ$, $\phi$: 41$^\circ$}}}} & 
    \includegraphics[width=\linewidth]{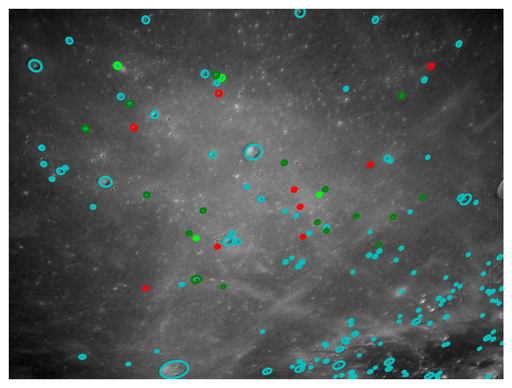} & 
    \includegraphics[width=\linewidth]{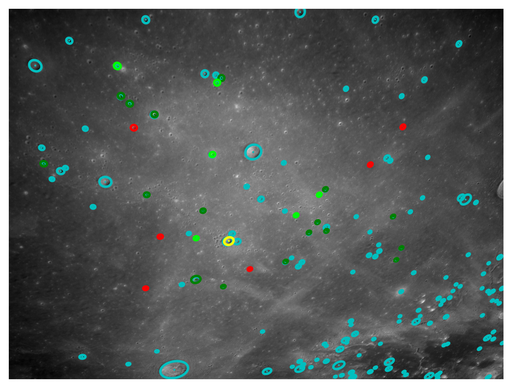} &
    \includegraphics[width=\linewidth]{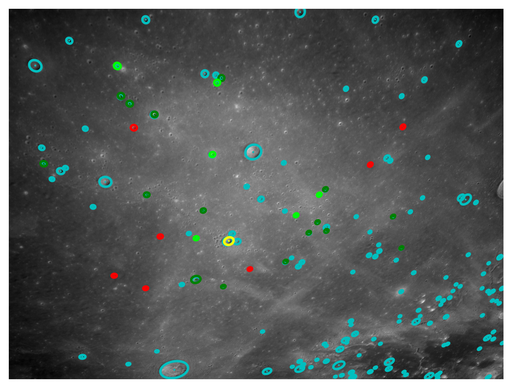} & 
    \includegraphics[width=\linewidth]{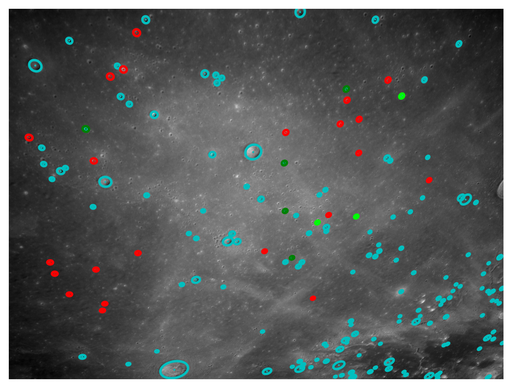} \\ 
    & 
    \includegraphics[width=.1\linewidth]{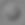} & 
    \includegraphics[width=.4\linewidth]{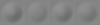} &
    \includegraphics[width=.8\linewidth]{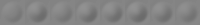} & 
    \includegraphics[width=.8\linewidth]{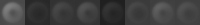} \\
    \parbox[t]{3mm}{\rotatebox[origin=l]{90}{\textcolor{gray}{\tiny{$\epsilon$: 17$^\circ$, $\iota$: 69$^\circ$, $\phi$: 53$^\circ$}}}} & 
    \includegraphics[width=\linewidth]{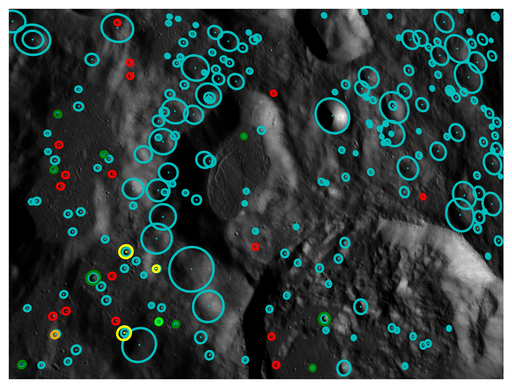} & 
    \includegraphics[width=\linewidth]{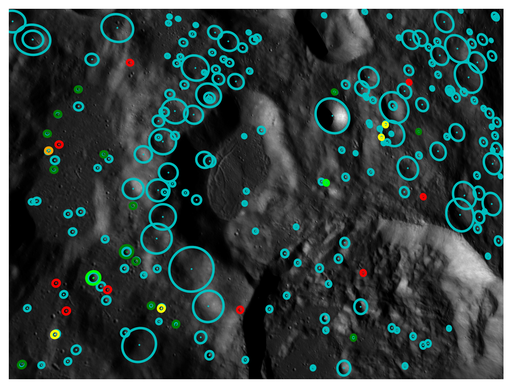} &
    \includegraphics[width=\linewidth]{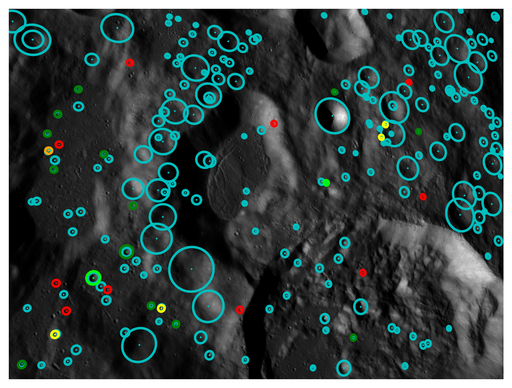} & 
    \includegraphics[width=\linewidth]{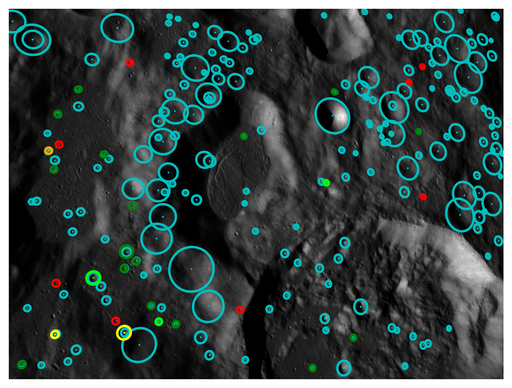} \\ 
    & 
    \includegraphics[width=.1\linewidth]{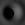} & 
    \includegraphics[width=.4\linewidth]{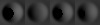} &
    \includegraphics[width=.8\linewidth]{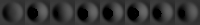} & 
    \includegraphics[width=.8\linewidth]{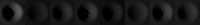} \\
    \parbox[t]{3mm}{\rotatebox[origin=l]{90}{\textcolor{gray}{\tiny{$\epsilon$: 46$^\circ$, $\iota$: 34$^\circ$, $\phi$: 67$^\circ$}}}} & 
    \includegraphics[width=\linewidth]{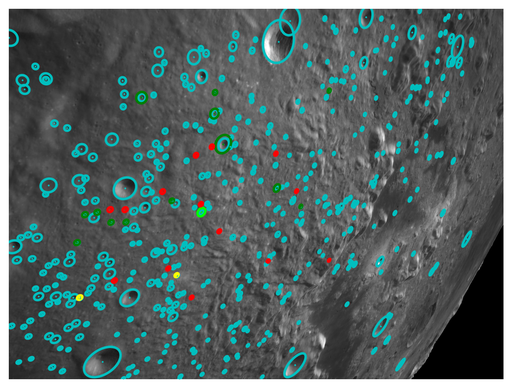} & 
    \includegraphics[width=\linewidth]{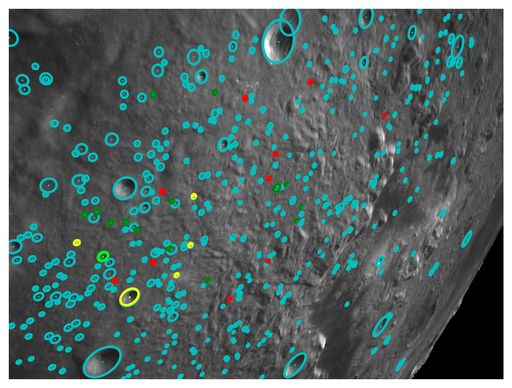} &
    \includegraphics[width=\linewidth]{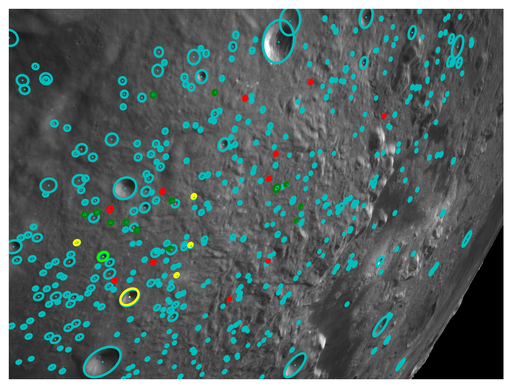} & 
    \includegraphics[width=\linewidth]{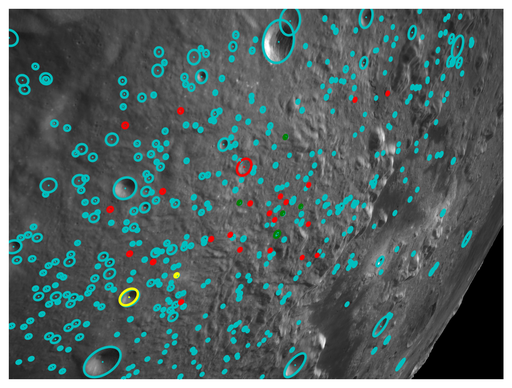} \\ 
    & 
    \includegraphics[width=.1\linewidth]{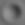} & 
    \includegraphics[width=.4\linewidth]{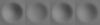} &
    \includegraphics[width=.8\linewidth]{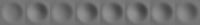} & 
    \includegraphics[width=.8\linewidth]{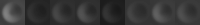} \\
    \parbox[t]{3mm}{\rotatebox[origin=l]{90}{\textcolor{gray}{\tiny{$\epsilon$: 6$^\circ$, $\iota$: 20$^\circ$, $\phi$: 19$^\circ$}}}} & 
    \includegraphics[width=\linewidth]{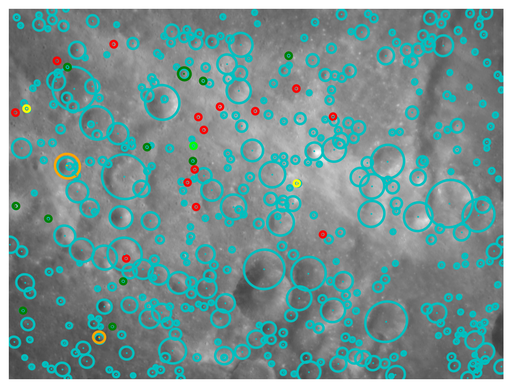} & 
    \includegraphics[width=\linewidth]{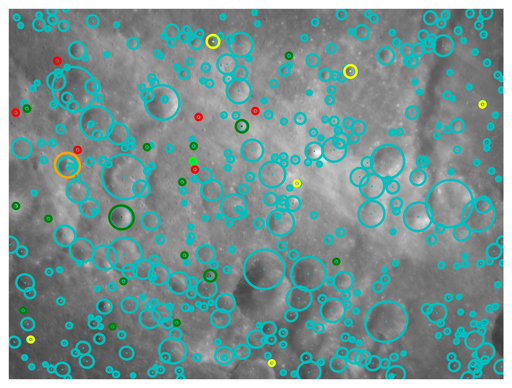} &
    \includegraphics[width=\linewidth]{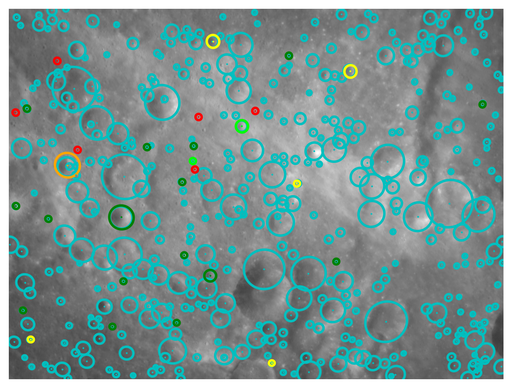} & 
    \includegraphics[width=\linewidth]{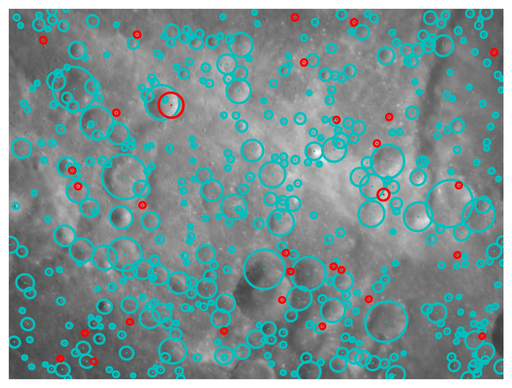} \\ 
    & 
    \includegraphics[width=.1\linewidth]{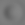} & 
    \includegraphics[width=.4\linewidth]{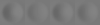} &
    \includegraphics[width=.8\linewidth]{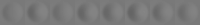} & 
    \includegraphics[width=.8\linewidth]{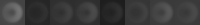} \\
    \parbox[t]{3mm}{\rotatebox[origin=l]{90}{\textcolor{gray}{\tiny{$\epsilon$: 29$^\circ$, $\iota$: 47$^\circ$, $\phi$: 18$^\circ$}}}} & 
    \includegraphics[width=\linewidth]{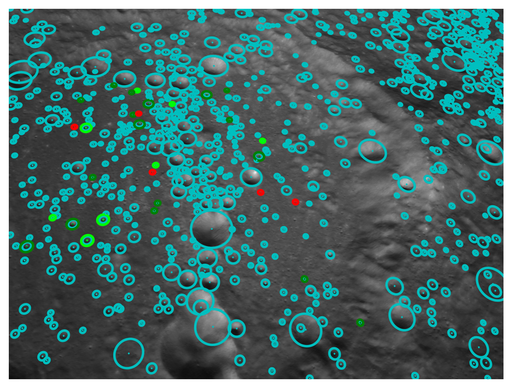} & 
    \includegraphics[width=\linewidth]{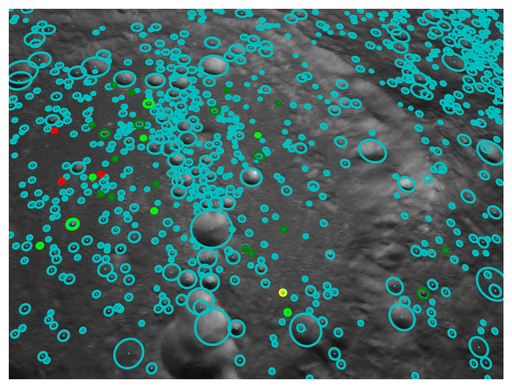} &
    \includegraphics[width=\linewidth]{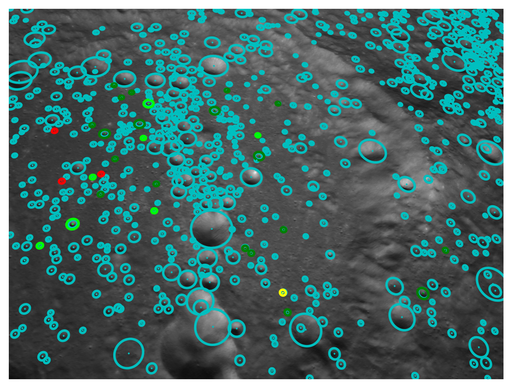} & 
    \includegraphics[width=\linewidth]{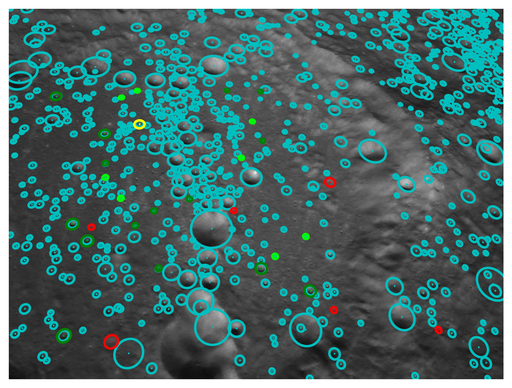} \\ 
    & 
    \includegraphics[width=.1\linewidth]{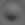} & 
    \includegraphics[width=.4\linewidth]{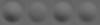} &
    \includegraphics[width=.8\linewidth]{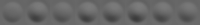} & 
    \includegraphics[width=.8\linewidth]{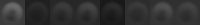} \\
    \parbox[t]{3mm}{\rotatebox[origin=l]{90}{\textcolor{gray}{\tiny{$\epsilon$: 53$^\circ$, $\iota$: 11$^\circ$, $\phi$: 64$^\circ$}}}} & 
    \includegraphics[width=\linewidth]{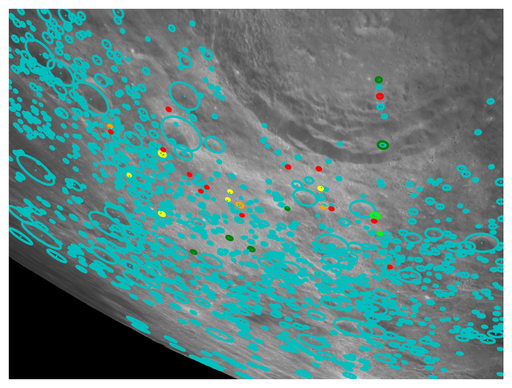} & 
    \includegraphics[width=\linewidth]{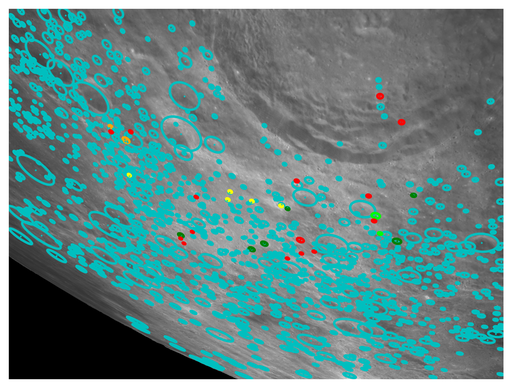} &
    \includegraphics[width=\linewidth]{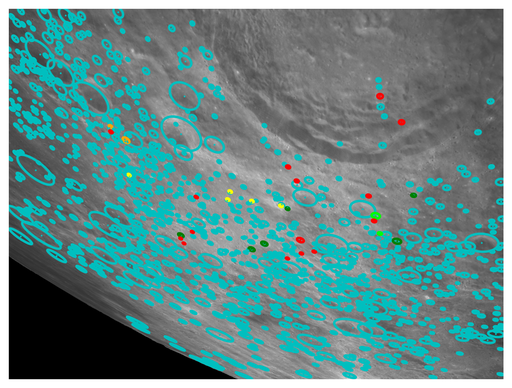} & 
    \includegraphics[width=\linewidth]{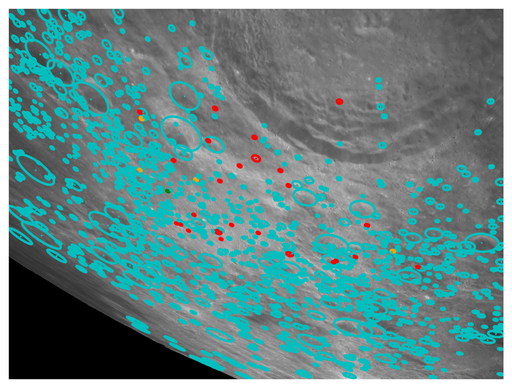} \\ 
    & 
    \includegraphics[width=.1\linewidth]{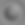} & 
    \includegraphics[width=.4\linewidth]{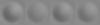} &
    \includegraphics[width=.8\linewidth]{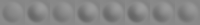} & 
    \includegraphics[width=.8\linewidth]{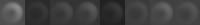} \\
\end{tabular}

    \end{adjustbox}
    \caption{\textbf{Qualitative comparison of detected craters for template matching methods.} Projected catalog craters are drawn in \textcolor{cyan}{\textbf{cyan}}, and detected craters are colored according to their distance to the nearest catalog crater with thresholds of \textcolor{mplLime}{$\mathbf{<1}$}, \textcolor{mplGreen}{$\mathbf{<3}$}, \textcolor{mplGold}{$\mathbf{<5}$}, \textcolor{mplOrange}{$\mathbf{<10}$}, and \textcolor{red}{$\mathbf{\geq 10}$} pixels. The emission ($\varepsilon$), incidence ($\iota$), and phase ($\phi$) angles for each image are provided in the left-most column.}
    \label{fig:detection-qual-compare}
\end{figure*}

Among the template-based approaches, the proposed \textsc{EigenCrater} method consistently outperforms the hand-designed Armstrong template~\cite{thrasher2024template} across all precision and recall thresholds. 
For example, \textsc{EigenCrater}4 achieves 60.49\% precision at 5 px compared to 48.83\% for the Armstrong template, with corresponding recall improvements from 3.48\% to 4.64\%. 
Increasing the number of clusters from four to eight provides virtually identical performance, suggesting diminishing returns beyond a modest number of representative templates. 
Incorporating albedo information does not consistently improve detection performance and, in most cases, reduces precision and recall, likely due to increased sensitivity to illumination and reflectance variability. 
Indeed, at low incidence and phase angles, albedo-driven intensity variations can dominate over topography-induced shading in the rendered templates, leading to false detections near non-crater surface highlights, as illustrated in Figure~\ref{fig:detection-qual-compare}.
Importantly, center localization errors for all template-based methods remain comparable or better to those of the fine-tuned Ellipse R-CNN, with sub-pixel accuracy at the strictest thresholds and errors below approximately 2.4 px at the 10 px tolerance. 
These results indicate that \textsc{EigenCrater} provides a substantial improvement over hand-selected templates while maintaining competitive localization accuracy, without requiring retraining or simulator-specific adaptation.

\begin{figure*}
    \centering
    \begin{adjustbox}{width=\linewidth}
    \centering
\setlength{\tabcolsep}{-1pt} %
\ra{0}
\begin{tabular}{c>{\centering\arraybackslash}p{3.3cm}>{\centering\arraybackslash}p{3.3cm}>{\centering\arraybackslash}p{3.3cm}>{\centering\arraybackslash}p{3.3cm}}
    & \scriptsize{Armstrong} & \scriptsize{\textsc{EigenCrater}4} & \scriptsize{Ellipse R-CNN (pretrained)} & \scriptsize{Ellipse R-CNN (finetuned)} \\
    \parbox[t]{3mm}{\rotatebox[origin=l]{90}{\textcolor{gray}{\tiny{$\epsilon$: 34$^\circ$, $\iota$: 7$^\circ$, $\phi$: 41$^\circ$}}}} & 
    \includegraphics[width=\linewidth]{figs/detections/armstrong/10001_nav_cam.png_armstrong.png} & 
    \includegraphics[width=\linewidth]{figs/detections/eigencrater4/10001_nav_cam.png_k4_elevonly.png} &
    \includegraphics[width=\linewidth]{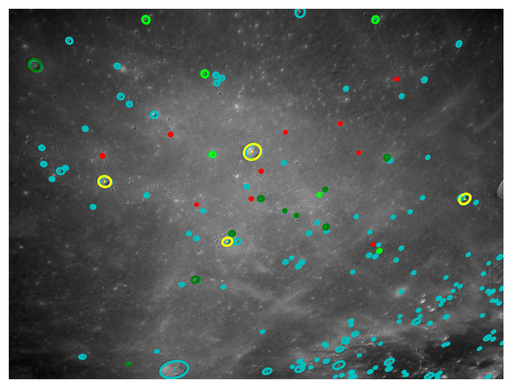} & 
    \includegraphics[width=\linewidth]{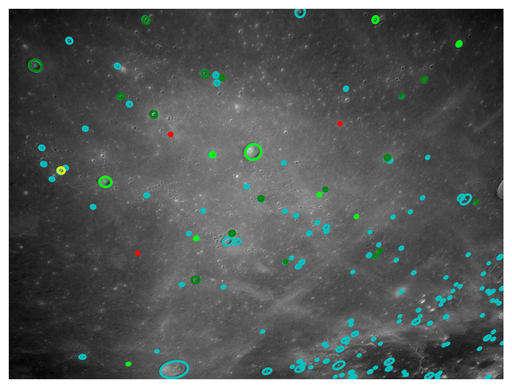} \\ 
    & 
    \includegraphics[width=.1\linewidth]{figs/detections/armstrong/10001_nav_cam_templates.png_armstrong.png} & 
    \includegraphics[width=.4\linewidth]{figs/detections/eigencrater4/10001_nav_cam_templates.png_k4_elevonly.png} \\
    \parbox[t]{3mm}{\rotatebox[origin=l]{90}{\textcolor{gray}{\tiny{$\epsilon$: 17$^\circ$, $\iota$: 69$^\circ$, $\phi$: 53$^\circ$}}}} & 
    \includegraphics[width=\linewidth]{figs/detections/armstrong/10017_nav_cam.png_armstrong.png} & 
    \includegraphics[width=\linewidth]{figs/detections/eigencrater4/10017_nav_cam.png_k4_elevonly.png} &
    \includegraphics[width=\linewidth]{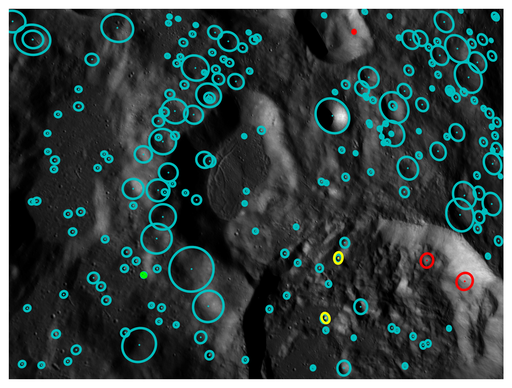} & 
    \includegraphics[width=\linewidth]{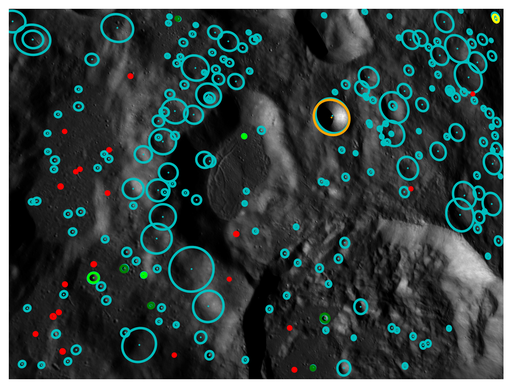} \\ 
    & 
    \includegraphics[width=.1\linewidth]{figs/detections/armstrong/10017_nav_cam_templates.png_armstrong.png} & 
    \includegraphics[width=.4\linewidth]{figs/detections/eigencrater4/10017_nav_cam_templates.png_k4_elevonly.png} \\
    \parbox[t]{3mm}{\rotatebox[origin=l]{90}{\textcolor{gray}{\tiny{$\epsilon$: 46$^\circ$, $\iota$: 34$^\circ$, $\phi$: 67$^\circ$}}}} & 
    \includegraphics[width=\linewidth]{figs/detections/armstrong/10019_nav_cam.png_armstrong.png} & 
    \includegraphics[width=\linewidth]{figs/detections/eigencrater4/10019_nav_cam.png_k4_elevonly.png} &
    \includegraphics[width=\linewidth]{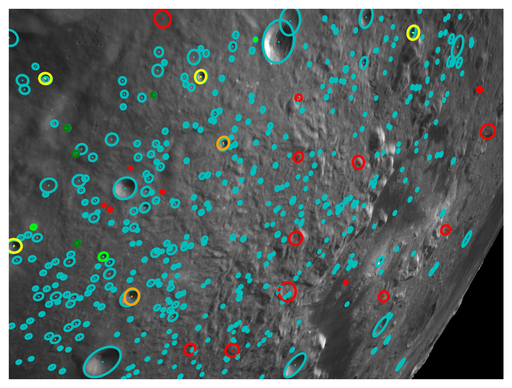} & 
    \includegraphics[width=\linewidth]{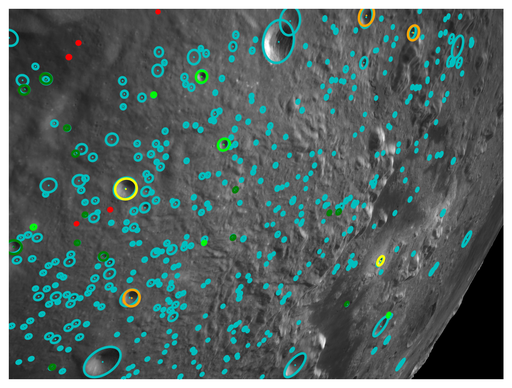} \\ 
    & 
    \includegraphics[width=.1\linewidth]{figs/detections/armstrong/10019_nav_cam_templates.png_armstrong.png} & 
    \includegraphics[width=.4\linewidth]{figs/detections/eigencrater4/10019_nav_cam_templates.png_k4_elevonly.png} \\
    \parbox[t]{3mm}{\rotatebox[origin=l]{90}{\textcolor{gray}{\tiny{$\epsilon$: 6$^\circ$, $\iota$: 20$^\circ$, $\phi$: 19$^\circ$}}}} & 
    \includegraphics[width=\linewidth]{figs/detections/armstrong/10026_nav_cam.png_armstrong.png} & 
    \includegraphics[width=\linewidth]{figs/detections/eigencrater4/10026_nav_cam.png_k4_elevonly.png} &
    \includegraphics[width=\linewidth]{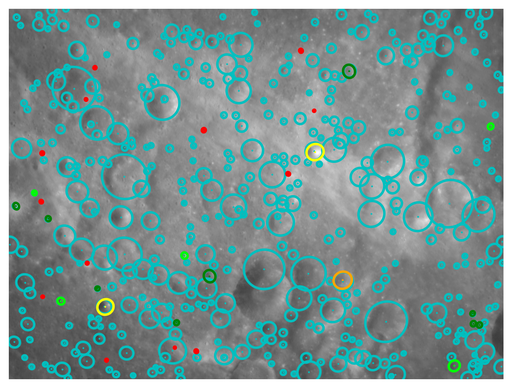} & 
    \includegraphics[width=\linewidth]{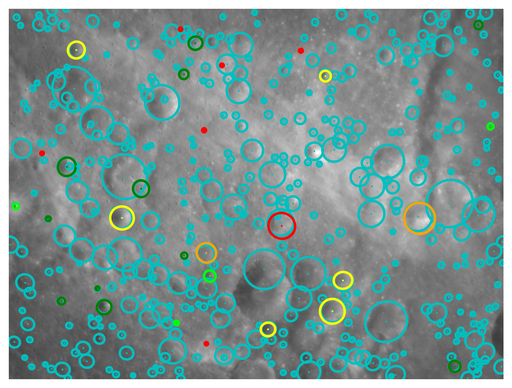} \\ 
    & 
    \includegraphics[width=.1\linewidth]{figs/detections/armstrong/10026_nav_cam_templates.png_armstrong.png} & 
    \includegraphics[width=.4\linewidth]{figs/detections/eigencrater4/10026_nav_cam_templates.png_k4_elevonly.png} \\
    \parbox[t]{3mm}{\rotatebox[origin=l]{90}{\textcolor{gray}{\tiny{$\epsilon$: 29$^\circ$, $\iota$: 47$^\circ$, $\phi$: 18$^\circ$}}}} & 
    \includegraphics[width=\linewidth]{figs/detections/armstrong/10037_nav_cam.png_armstrong.png} & 
    \includegraphics[width=\linewidth]{figs/detections/eigencrater4/10037_nav_cam.png_k4_elevonly.png} &
    \includegraphics[width=\linewidth]{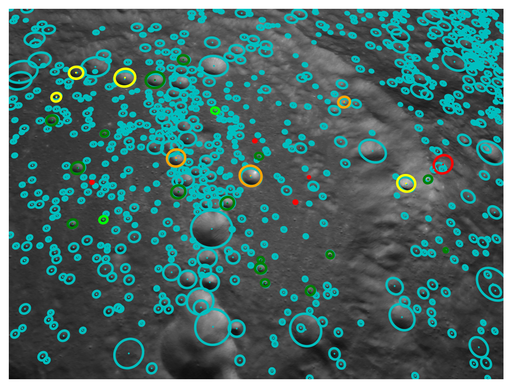} & 
    \includegraphics[width=\linewidth]{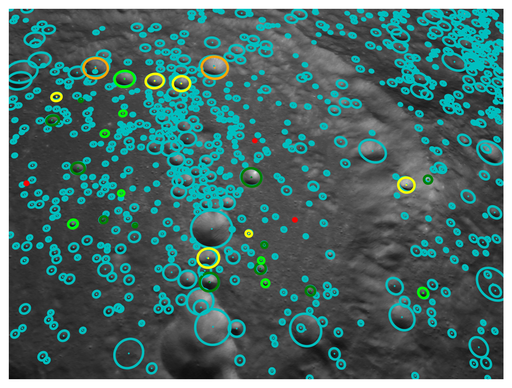} \\ 
    & 
    \includegraphics[width=.1\linewidth]{figs/detections/armstrong/10037_nav_cam_templates.png_armstrong.png} & 
    \includegraphics[width=.4\linewidth]{figs/detections/eigencrater4/10037_nav_cam_templates.png_k4_elevonly.png} \\
    \parbox[t]{3mm}{\rotatebox[origin=l]{90}{\textcolor{gray}{\tiny{$\epsilon$: 53$^\circ$, $\iota$: 11$^\circ$, $\phi$: 64$^\circ$}}}} & 
    \includegraphics[width=\linewidth]{figs/detections/armstrong/10043_nav_cam.png_armstrong.png} & 
    \includegraphics[width=\linewidth]{figs/detections/eigencrater4/10043_nav_cam.png_k4_elevonly.png} &
    \includegraphics[width=\linewidth]{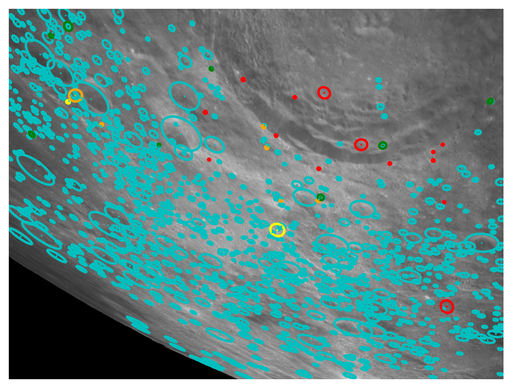} & 
    \includegraphics[width=\linewidth]{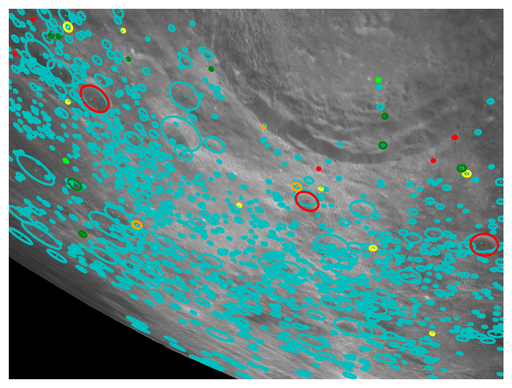} \\ 
    & 
    \includegraphics[width=.1\linewidth]{figs/detections/armstrong/10043_nav_cam_templates.png_armstrong.png} & 
    \includegraphics[width=.4\linewidth]{figs/detections/eigencrater4/10043_nav_cam_templates.png_k4_elevonly.png} \\
\end{tabular}

    \end{adjustbox}
    \caption{\textbf{Qualitative comparison of detected craters for template matching and Ellipse R-CNN.} Projected catalog craters are drawn in \textcolor{cyan}{\textbf{cyan}}, and detected craters are colored according to their distance to the nearest catalog crater with thresholds of \textcolor{mplLime}{$\mathbf{<1}$}, \textcolor{mplGreen}{$\mathbf{<3}$}, \textcolor{mplGold}{$\mathbf{<5}$}, \textcolor{mplOrange}{$\mathbf{<10}$}, and \textcolor{red}{$\mathbf{\geq 10}$} pixels. The emission ($\varepsilon$), incidence ($\iota$), and phase ($\phi$) angles for each image are provided in the left-most column.}
    \label{fig:detection-qual-compare-with-ercnn}
\end{figure*}

\begin{figure*}[p]
    \centering
    \begin{subfigure}{\linewidth}
        \includegraphics[width=\linewidth]{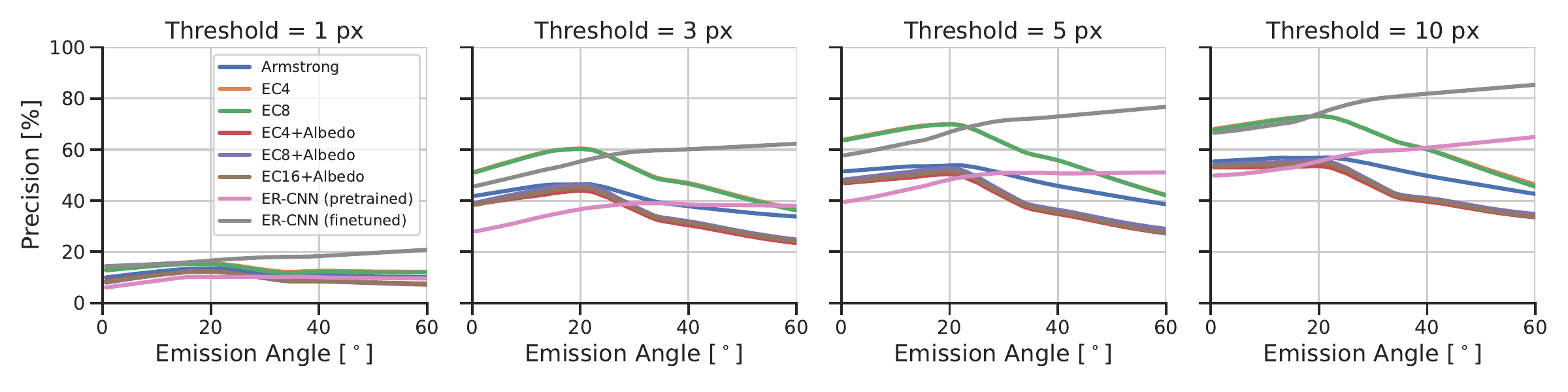}
    \end{subfigure}\\
    \begin{subfigure}{\linewidth}
        \includegraphics[width=\linewidth]{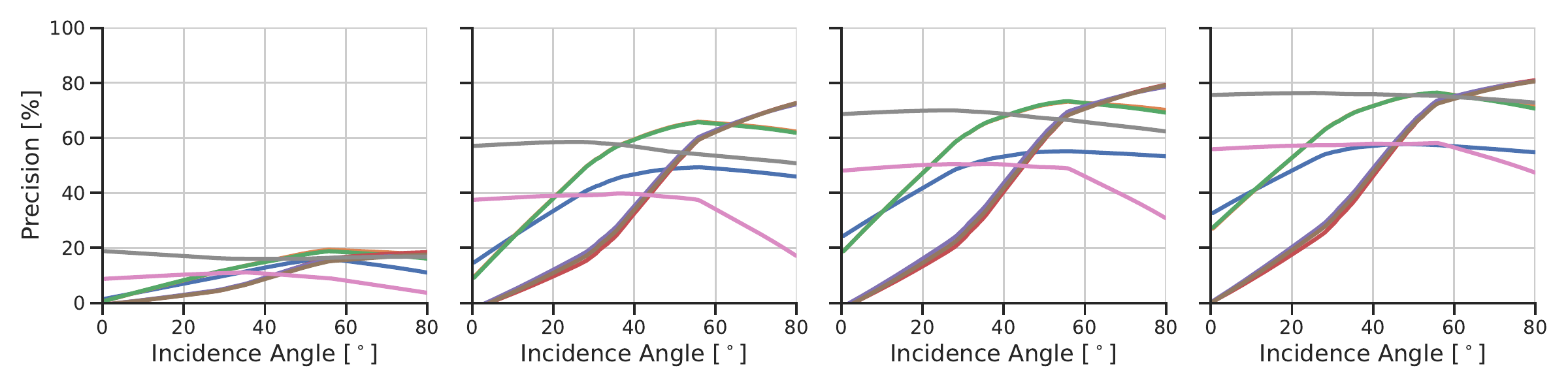}
    \end{subfigure}\\
    \begin{subfigure}{\linewidth}
        \includegraphics[width=\linewidth]{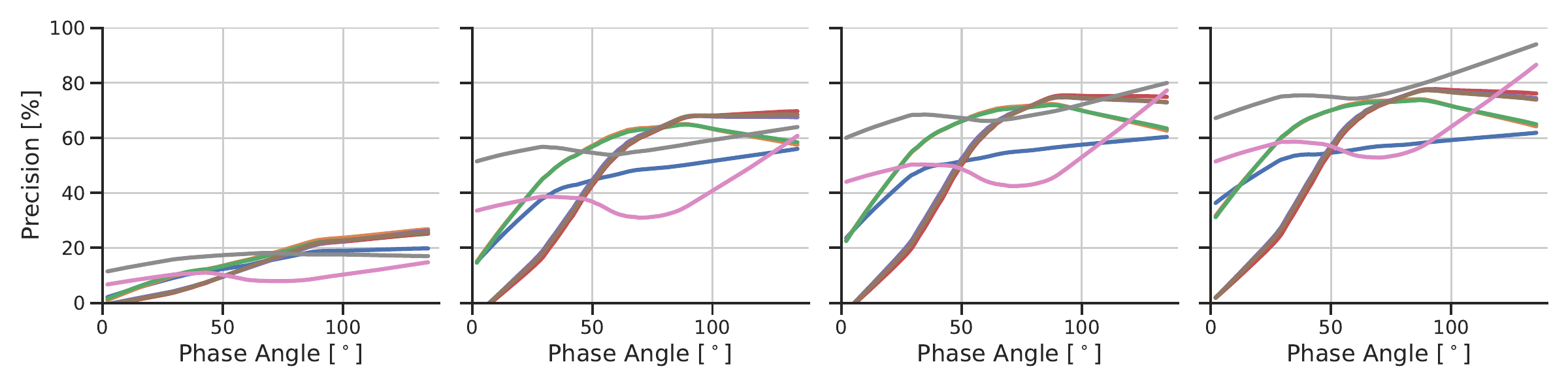}
    \end{subfigure}
    \caption{\textbf{Detection precision versus illumination and viewing geometry.} Trends are visualized using locally weighted scatterplot smoothing (LOWESS)~\cite{cleveland1979lowess,seabold2010statsmodels}.}
    \label{fig:precision-versus-geometry}
\end{figure*}

Figure~\ref{fig:precision-versus-geometry} illustrates detection precision as a function of illumination and viewing geometry. 
The fine-tuned Ellipse R-CNN model maintains strong performance across a broad range of incidence and emission angles, whereas the pretrained model exhibits significantly lower precision, particularly at higher incidence angles. 
\textsc{EigenCrater} performs exceptionally well at emission angles below 40$^\circ$ and incidence angles above 20$^\circ$, even outperforming the fine-tuned Ellipse R-CNN model over some regions. 
However, performance degrades near opposition (low phase angles) and at large off-nadir viewing angles, where shading cues become less distinctive and geometric distortion increases. 
As discussed previously, incorporating albedo information further reduces precision at low incidence and phase angles due to the dominance of albedo-driven intensity variations, which do not seem to produce reliable detections. 
These trends indicate that crater-based template matching performs most reliably under moderate illumination conditions and near-nadir viewing geometries, highlighting the importance of mission planning to ensure favorable viewing and lighting configurations when employing crater templates for optical navigation.

\subsection{Position Estimation Performance}

\begin{figure*}[p]
    \centering
    \includegraphics[width=\linewidth]{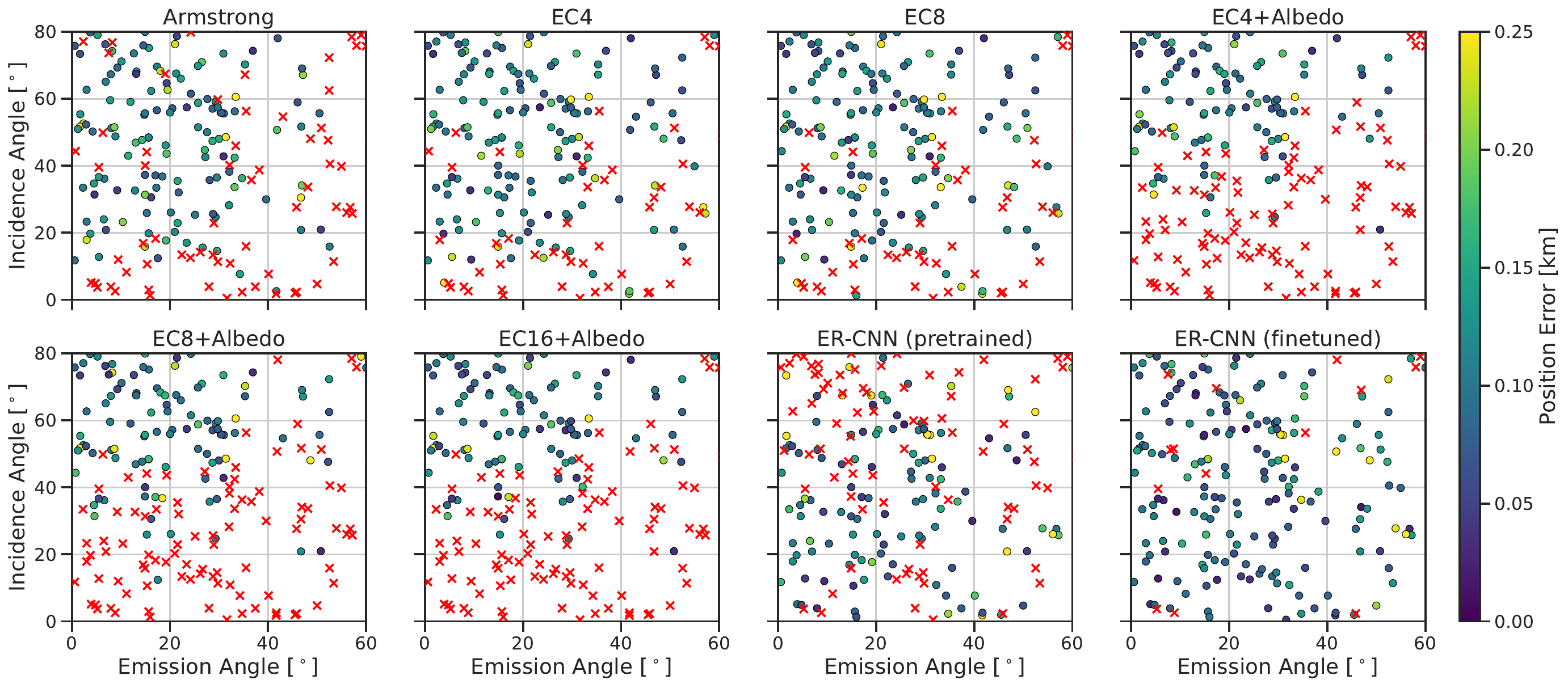}
    \caption{\textbf{Position error versus illumination and viewing geometry.} \textcolor{red}{$\mathbf{\times}$}'s indicate instances when position estimation failed due to insufficient inliers after the RANSAC verification step.}
    \label{fig:position-error-emission-incidence-scatter}
\end{figure*}

\begin{table}[tb!]
    \centering
    \ra{1.5}
    \caption{\textbf{Position estimation statistics.} The success rate (SR) quantifies the percentage of images where position estimation was successful. See the \nameref{sec:metrics} section for metric definitions. }
    \begin{adjustbox}{width=\linewidth}
    \begin{tabular}{lrrrrrrrrr}
\toprule
&& \multicolumn{3}{c}{Position Error [km]} && \multicolumn{4}{c}{AUC} \\
\cmidrule{3-5} \cmidrule{7-10}
 & SR [\%] & Mean $\pm$ $1\sigma$ & Min. & Max && @ 0.1 km & @ 0.3 km & @ 0.5 km & @ 1 km \\ 
\midrule
Ellipse R-CNN~\cite{dong2021ellipse,doppenberg2021ercnn} (pretrained) & 61.5 & $0.13\pm 0.10$ & 0.02 & 0.60 && 10.65 &     36.84 &     45.99 &   53.74 \\
Ellipse R-CNN~\cite{dong2021ellipse,doppenberg2021ercnn} (finetuned)  & \textbf{93.0} & \textbf{0.10} $\pm$ \textbf{0.06} & \textbf{0.01} & \textbf{0.34} && \textbf{19.82} &     \textbf{62.04} &     \textbf{74.41} &   \textbf{83.71} \\
\midrule
Armstrong~\cite{thrasher2024template} & 66.5 & 0.13 $\pm$ 0.07 & 0.03 & 0.57 &&  6.96 &     39.48 &     49.89 &   58.23 \\
\textsc{EigenCrater}4                 & \underline{75.5} & \underline{0.12} $\pm$ \underline{0.06} & \underline{0.02} & 0.54 &&  \underline{8.88} &     \underline{45.55} &     \underline{57.33} &   \underline{66.46} \\
\textsc{EigenCrater}4$+$Albedo        & 53.0 & 0.12 $\pm$ 0.08 & 0.03 & \underline{0.52} &&  7.49 &     32.71 &     40.49 &   46.75 \\
\textsc{EigenCrater}8                 & \textbf{77.5} & 0.12 $\pm$ 0.07 & \underline{0.02} & 0.59 &&  \textbf{9.21} &     \textbf{46.32} &     \textbf{58.45} &   \textbf{67.98} \\
\textsc{EigenCrater}8$+$Albedo        & 53.0 & 0.12 $\pm$ 0.08 & 0.03 & \underline{0.52} &&  7.49 &     32.71 &     40.49 &   46.75 \\
\textsc{EigenCrater}16$+$Albedo       & 51.0 & \textbf{0.11} $\pm$ \textbf{0.06} & \textbf{0.01} & \textbf{0.48} &&  8.75 &     33.43 &     40.39 &   45.70 \\
\bottomrule
\end{tabular}

    \end{adjustbox}
    \label{tab:position-estimation-results}
\end{table}

The position estimation results in Table~\ref{tab:position-estimation-results} exhibit trends consistent with the detection performance observed previously. 
Fine-tuning Ellipse R-CNN yields the strongest overall navigation performance, achieving a 93.0\% success rate (SR), the lowest mean position error of $0.10 \pm 0.06$ km, and the highest position error AUC values across all thresholds. 
In contrast, the pretrained Ellipse R-CNN model achieves a substantially lower SR of 61.5\%, despite a comparable mean error among successful cases. 
This gap further illustrates the sensitivity of learning-based pipelines to domain alignment.

Among the template-based approaches, \textsc{EigenCrater}4 and \textsc{EigenCrater}8 provide a clear improvement over the Armstrong template in terms of success rate and AUC, achieving SR values of 75.5\% and 77.5\%, respectively, compared to 66.5\% for Armstrong. 
Although the mean position errors for successful runs remain similar across template-based methods (approximately $0.12 \pm 0.06$ km), the higher success rate and improved AUC indicate significantly higher position estimation recall. 
Increasing the number of clusters from four to eight yields only marginal gains in AUC, again suggesting diminishing returns beyond a modest template set. 
Incorporating albedo information reduces the success rate and AUC across all configurations, despite occasionally lowering the mean error among successful cases, reinforcing the earlier observation that albedo-driven intensity variations can introduce spurious detections under certain illumination conditions. 
Overall, these results demonstrate that \textsc{EigenCrater} improves navigation reliability relative to hand-selected templates while maintaining competitive localization accuracy, providing a strong balance between robustness and interpretability without requiring retraining on mission-specific imagery.

Figure~\ref{fig:position-error-emission-incidence-scatter} illustrates position estimation error as a function of emission and incidence angles. 
No strong trends are observed between position error and viewing or illumination geometry among successful solutions. 
However, \textsc{EigenCrater}4 and \textsc{EigenCrater}8 fail to reliably estimate the camera's position at incidence angles of less than 20$^\circ$ and emission angles of greater than 50$^\circ$. 
Incorporating albedo information further reduces position estimation recall at incidence angles below 40$^\circ$, consistent with the previously discussed sensitivity to albedo-dominated intensity variations under low-incidence conditions. 
The pretrained Ellipse R-CNN model similarly struggles at high incidence angles (greater than 60$^\circ$) and larger emission angles (greater than 40$^\circ$), whereas the fine-tuned model maintains reliable position estimation across nearly the full range of viewing and illumination geometries considered.


\subsection{Homographic Adaptation Ablation Study} \label{sec:homography-ablation}

\begin{figure}[tb!]
    \centering
    \begin{adjustbox}{width=\linewidth}
    \centering
\setlength{\tabcolsep}{-1pt} %
\ra{0}
\begin{tabular}{c>{\centering\arraybackslash}p{3.3cm}>{\centering\arraybackslash}p{3.3cm}>{\centering\arraybackslash}p{3.3cm}>{\centering\arraybackslash}p{3.3cm}}
    & \scriptsize{No transformation} & \scriptsize{No warp} & \scriptsize{Warp template} & \scriptsize{Warp image} \\
    \parbox[t]{3mm}{\rotatebox[origin=l]{90}{\textcolor{gray}{\tiny{$\epsilon$: 34$^\circ$, $\iota$: 7$^\circ$, $\phi$: 41$^\circ$}}}} & 
    \includegraphics[width=\linewidth]{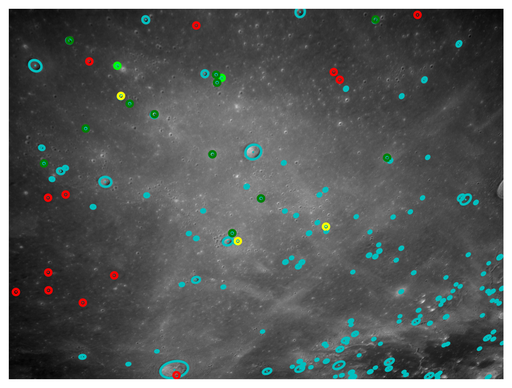} &
    \includegraphics[width=\linewidth]{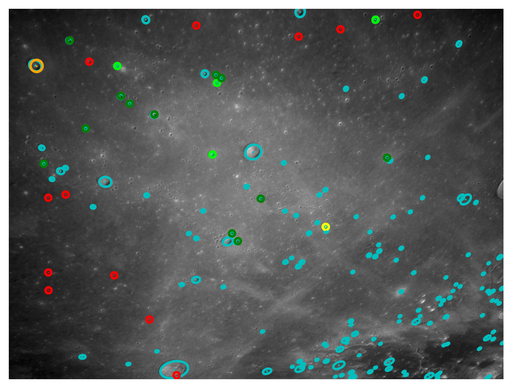} &
    \includegraphics[width=\linewidth]{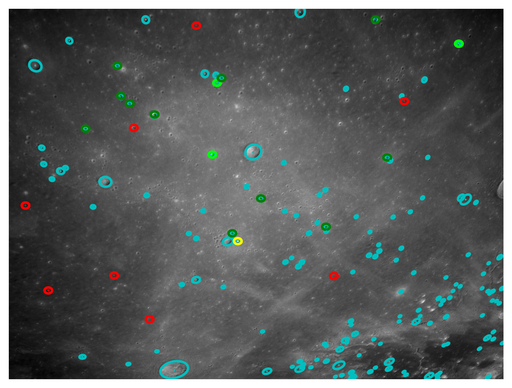} & 
    \includegraphics[width=\linewidth]{figs/detections/eigencrater4/10001_nav_cam.png_k4_elevonly.png} \\
    & 
    \includegraphics[height=.1\linewidth]{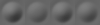} &
    \includegraphics[height=.1\linewidth]{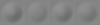} &
    \includegraphics[height=.1\linewidth]{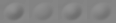} & 
    \includegraphics[height=.1\linewidth]{figs/detections/eigencrater4/10001_nav_cam_templates.png_k4_elevonly.png} \\
    \parbox[t]{3mm}{\rotatebox[origin=l]{90}{\textcolor{gray}{\tiny{$\epsilon$: 34$^\circ$, $\iota$: 7$^\circ$, $\phi$: 41$^\circ$}}}} & 
    \includegraphics[width=\linewidth]{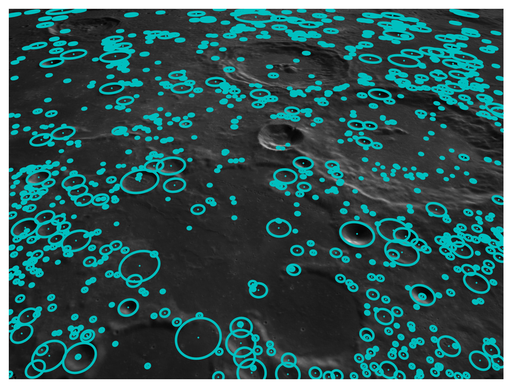} &
    \includegraphics[width=\linewidth]{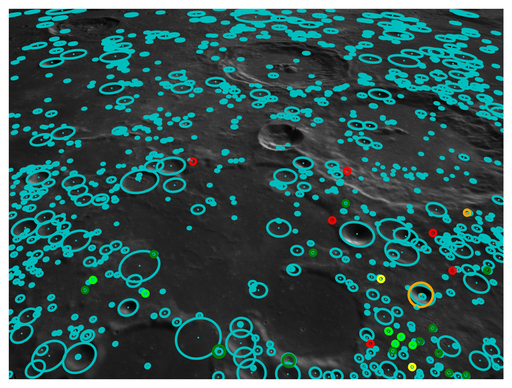} &
    \includegraphics[width=\linewidth]{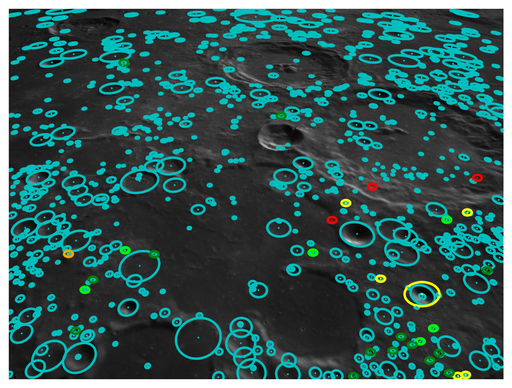} & 
    \includegraphics[width=\linewidth]{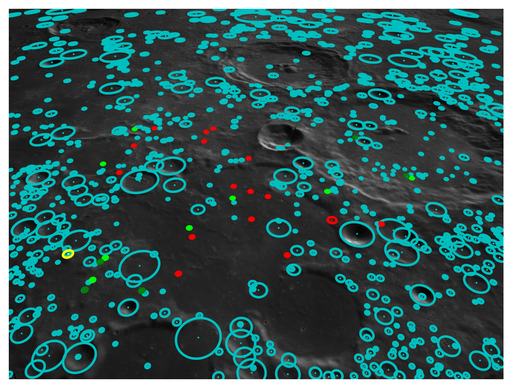} \\
    & 
    \includegraphics[height=.1\linewidth]{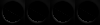} &
    \includegraphics[height=.1\linewidth]{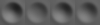} &
    \includegraphics[height=.1\linewidth]{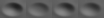} & 
    \includegraphics[height=.1\linewidth]{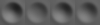} \\
\end{tabular}

    \end{adjustbox}
    \caption{\textbf{Qualitative comparison of detected craters for each warping procedure.} Projected catalog craters are drawn in \textcolor{cyan}{\textbf{cyan}}, and detected craters are colored according to their distance to the nearest catalog crater with thresholds of \textcolor{mplLime}{$\mathbf{<1}$}, \textcolor{mplGreen}{$\mathbf{<3}$}, \textcolor{mplGold}{$\mathbf{<5}$}, \textcolor{mplOrange}{$\mathbf{<10}$}, and \textcolor{red}{$\mathbf{\geq 10}$} pixels. The emission ($\varepsilon$), incidence ($\iota$), and phase ($\phi$) angles for each image are provided in the left-most column.}
    \label{fig:homography-qual-compare}
\end{figure}

Table~\ref{tab:homography-adaptation-ablation} summarizes the impact of homographic adaptation for compensating off-nadir viewing geometry with qualitative examples provided in Figure~\ref{fig:homography-qual-compare}. 
Applying the homography directly to the input image yields the strongest overall performance, achieving the highest precision (52.45\%), recall (4.09\%), and lowest center error (1.56 px), along with the best AUC values at the tighter position error thresholds (0.1 km, 0.3 km, and 0.5 km). 
Warping the template instead provides only marginal improvements in precision over the unwarped case and results in degraded AUC, suggesting that distortions introduced during template resampling may reduce correlation fidelity and recall. 
The ``No transformation'' case corresponds no warping and not transforming the incidence and emission vectors into the computed nadir frame. 
Intuitively, this results in rendering that do not match the image, especially at large emission angles, leading to the lowest detection and position estimation perform. 

\begin{table*}[tb!]
    \centering
    \ra{1.5}
    \caption{\textbf{Homographic adaptation ablation for \textsc{EigenCrater}4.} Detection precision (P), recall (R), and center errors (CE) are reported at a 3 pixel threshold.}
    \begin{adjustbox}{width=0.75\linewidth}
    \begin{tabular}{lrrrrrrr}
\toprule
& \multicolumn{3}{c}{} & \multicolumn{4}{c}{Position Error AUC} \\
 \cmidrule{5-8}
 & P [\%] & R [\%] & CE [px] & @ 0.1 km & @ 0.3 km & @ 0.5 km & @ 1 km \\ 
\midrule
No transformation & 42.70 & 3.28 & 1.68 & 5.79 &     37.57 &     49.15 &   58.54 \\
No warping        & 49.12 & \underline{3.80} & \underline{1.61} & \underline{7.05} &     \underline{43.14} &     \underline{56.22} &   \underline{66.86} \\
Warp template     & \underline{50.44} & 3.69 & 1.65 & 6.52 &     41.98 &     55.86 &   \textbf{66.88} \\
Warp image        & \textbf{52.45} & \textbf{4.09} & \textbf{1.56} & \textbf{8.88} &     \textbf{45.55} &     \textbf{57.33} &   66.46 \\
\bottomrule
\end{tabular}

    \end{adjustbox}
    \label{tab:homography-adaptation-ablation}
\end{table*}

\section{Conclusion}

We believe \textsc{EigenCrater} is an attractive alternative to deep learning methods for missions that require high interpretability, such as human space flight missions. 
While methods such as Ellipse R-CNN provide precise and reliable crater detections, it is essential to carefully characterize the sim-to-real gap (or sim-to-sim gap in our experiments) before deploying learning-based systems in safety-critical applications. 
In contrast, \textsc{EigenCrater} relies on physically grounded elevation data and transparent template correlation, yielding predictable behavior and well-understood failure modes without requiring mission-specific retraining. 
Although learning-based approaches may ultimately achieve superior performance under domain-matched conditions, the proposed method offers a compelling balance between robustness, interpretability, and implementation simplicity, making it well suited for risk-sensitive optical navigation architectures.

\section*{Funding Sources}
A portion of this work was supported by NASA cooperative agreement 80NSSC22M0151. 

\section*{Acknowledgments}

The authors thank Ava Thrasher for several helpful discussions on template-based crater detection, and Ellis King and Daniel Posada for their constructive technical insights and valuable feedback.

\bibliographystyle{AAS_publication}   
\bibliography{references}

\end{document}